\let\mypdfximage\pdfximage
\def\pdfximage{\immediate\mypdfximage}

\documentclass[runningheads]{llncs}
\usepackage{graphicx}
\usepackage{amsmath,amssymb} 

\usepackage{color}
\usepackage[width=122mm,left=12mm,paperwidth=146mm,height=193mm,top=12mm,paperheight=217mm]{geometry}
\usepackage{amsmath,amssymb} 
\usepackage{color}
\usepackage{caption}
\usepackage{subcaption}
\usepackage{url}
\usepackage{xcolor}
\usepackage{rotating}
\usepackage[bookmarks=false, colorlinks=true, pdfborder={0,0,0}, linkbordercolor={1 1 1}]{hyperref}
\usepackage{booktabs}

\begin{document}
\pagestyle{headings}
\mainmatter
\def\ECCV18SubNumber{2775}  

\title{OmniDepth: Dense Depth Estimation for Indoors Spherical Panoramas.}

\titlerunning{OmniDepth}

\author{Nikolaos Zioulis\inst{1}\thanks{Indicates equal contribution.}\and Antonis Karakottas\inst{1}\protect\footnotemark[1]\and Dimitrios Zarpalas\inst{1}\and Petros Daras\inst{1}}
\authorrunning{Zioulis et al.}

\institute{Centre for Research and Technology Hellas (CERTH) - Information Technologies Institute (ITI) - Visual Computing Lab (VCL), Thessaloniki, Greece\\ \email{\{nzioulis, ankarako, zarpalas, daras\}@iti.gr} \\ \url{vcl.iti.gr}}

\def\Dataset{360D }
\def\DATASET{\textbf{\Dataset Dataset: }}
\newcommand{\degree}[1]{${#1}^o$}
\newcommand{\conv}[2]{${#1} \times {#2}$}
\def\360{\degree{360}}
\def\unbal{UResNet }
\def\unbalnospace{UResNet}
\def\rectdil{RectNet }
\def\rectdilnospace{RectNet}
\def\UNBAL{\textbf{UResNet: }}
\def\RECTDIL{\textbf{RectNet: }}
\def\datasetURL{http://vcl.iti.gr/360-dataset/}
\maketitle

\begin{abstract}
Recent work on depth estimation up to now has only focused on projective images ignoring \360 content which is now increasingly and more easily produced. We show that monocular depth estimation models trained on traditional images produce sub-optimal results on omnidirectional images, showcasing the need for training directly on \360 datasets, which however, are hard to acquire. In this work, we circumvent the challenges associated with acquiring high quality \360 datasets with ground truth depth annotations, by re-using recently released large scale 3D datasets and re-purposing them to \360 via rendering. This dataset, which is considerably larger than similar projective datasets, is publicly offered to the community to enable future research in this direction. We use this dataset to learn in an end-to-end fashion the task of depth estimation from \360 images. We show promising results in our synthesized data as well as in unseen realistic images.
\end{abstract}

\section{Introduction}
\label{sec:intro}
One of the fundamental challenges in computer and 3D vision is the estimation of a scene's depth. Depth estimation leads to a three-dimensional understanding of the world which is very important to numerous applications. These vary from creating 3D maps \cite{tateno2017cnnslam} and allowing navigation in real-world environments \cite{Mo18AdobeIndoorNav}, to enabling stereoscopic rendering \cite{hedman2017casual}, synthesizing novel views out of pre-captured content \cite{huang20176} and even compositing 3D objects into it \cite{karsch2014automatic}. Depth information has been shown to boost the effectiveness of many vision tasks related to scene understanding when utilized jointly with color information \cite{eigen2015predicting,ren2012rgb}.

\begin{figure}[!ht]
	\centering
\resizebox{\textwidth}{!}{
	\begin{subfigure}[t]{0.3\textwidth}
    \centering
    	\includegraphics[scale = 0.2]{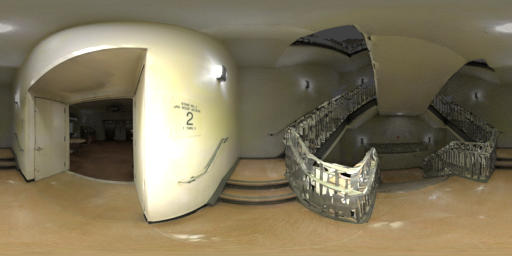}
    \end{subfigure}
    \begin{subfigure}[t]{0.3\textwidth}
    \centering
    	\includegraphics[scale = 0.2]{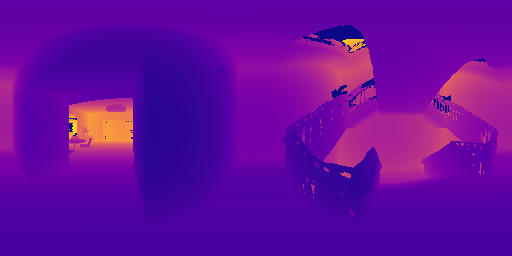}
    \end{subfigure}
    \begin{subfigure}[t]{0.3\textwidth}
    \centering
    	\includegraphics[scale = 0.2]{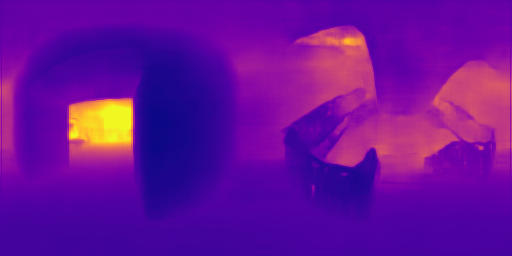}
    \end{subfigure} \rule{0.05cm}{2cm}
	\begin{subfigure}[t]{0.3\textwidth}
		\centering
		\includegraphics[scale = 0.2]{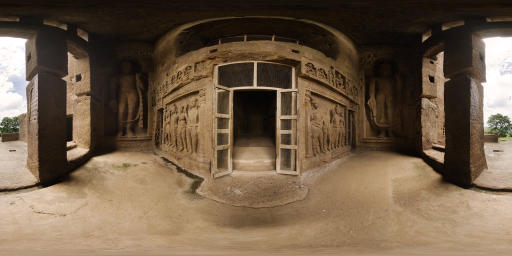}
	\end{subfigure} 
	\begin{subfigure}[t]{0.3\textwidth}
		\centering
		\includegraphics[scale = 0.2]{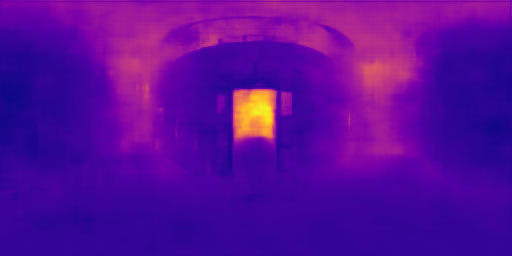}
	\end{subfigure} 	
}

\resizebox{\textwidth}{!}{
	\begin{subfigure}[t]{0.3\textwidth}
		\centering
		\includegraphics[scale = 0.2]{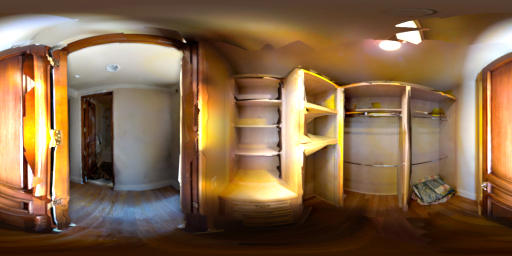}
	\end{subfigure}
	\begin{subfigure}[t]{0.3\textwidth}
		\centering
		\includegraphics[scale = 0.2]{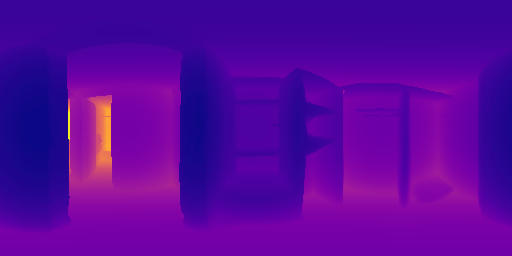}
	\end{subfigure}
	\begin{subfigure}[t]{0.3\textwidth}
		\centering
		\includegraphics[scale = 0.2]{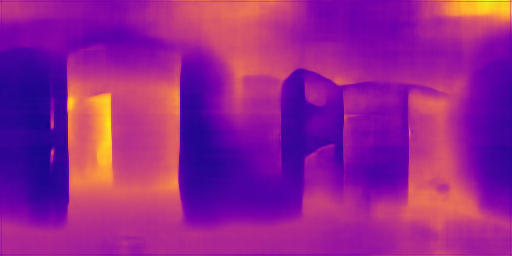}
	\end{subfigure} \rule{0.05cm}{2cm}
	\begin{subfigure}[t]{0.3\textwidth}
		\centering
		\includegraphics[scale = 0.2]{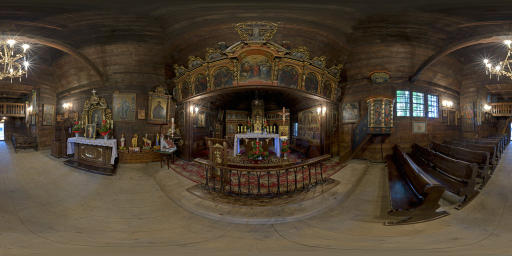}
	\end{subfigure} 
	\begin{subfigure}[t]{0.3\textwidth}
		\centering
		\includegraphics[scale = 0.2]{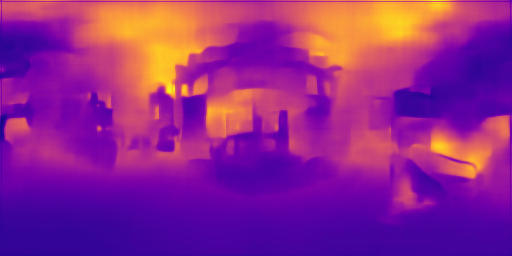}
	\end{subfigure} 	
}

\resizebox{\textwidth}{!}{
	\begin{subfigure}[b]{0.3\textwidth}
		\centering
		\includegraphics[scale = 0.2]{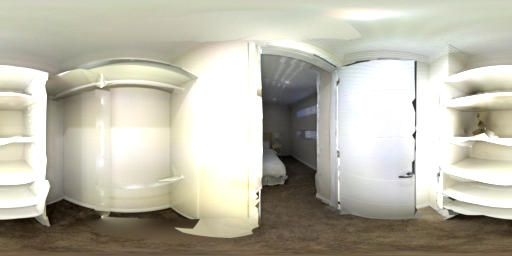}
		\caption{}
	\end{subfigure}
	\begin{subfigure}[b]{0.3\textwidth}
		\centering
		\includegraphics[scale = 0.2]{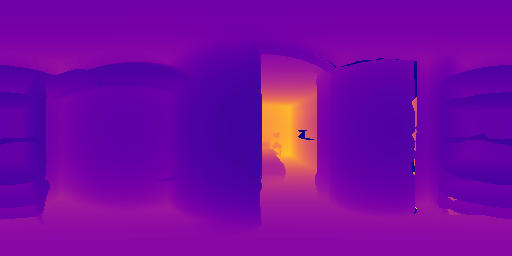}
		\caption{}
	\end{subfigure}
	\begin{subfigure}[b]{0.3\textwidth}
		\centering
		\includegraphics[scale = 0.2]{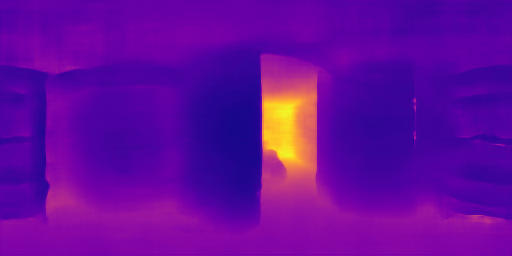}
		\caption{}
	\end{subfigure} \rule[2mm]{0.05cm}{2.3cm}
	\begin{subfigure}[b]{0.3\textwidth}
		\centering
		\includegraphics[scale = 0.83]{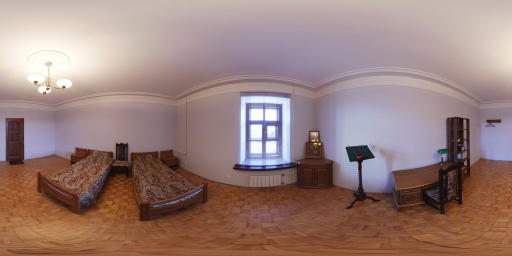}
		\caption{}
	\end{subfigure}
	\begin{subfigure}[b]{0.3\textwidth}
		\centering
		\includegraphics[scale = 0.2]{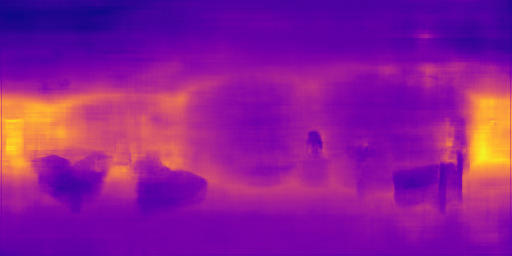}
		\caption{}
	\end{subfigure}
}
\caption{We learn the task of predicting depth directly from omnidirectional indoor scene images. Results from our \rectdil model are presented (left to right): (a) \360 image samples from our test set, (b) corresponding ground truth depth, (c) predicted depth maps of the test image samples, (d) \360 unseen image samples from the Sun360 dataset, (e) predicted depth maps of the Sun360 image samples.}
\label{fig:intro}
\end{figure}

Similar to how babies start to perceive depth from two viewpoints and then by ego-motion and observation of objects' motions, researchers have tackled the problem of estimating depth via methods built on multi-view consistency \cite{hartley2000multiple,furukawa2015multi} and structure-from-motion (SfM)\cite{ozyecsil2017survey}. But humans are also driven by past experiences and contextual similarities and apply this collective knowledge when presented with new scenes. Likewise, with the advent of more effective machine learning techniques, recent research focuses on learning to predict depth and has led to impressive results even with completely unsupervised learning approaches. 

However, learning based approaches have only focused on traditional 2D content captured by typical pinhole projection model based cameras. With the emergence of efficient spherical cameras and rigs, omnidirectional (\360) content is now more easily and consistently produced and is witnessing increased adoption in entertainment and marketing productions, robotics and vehicular applications as well as coverage of events and even journalism. Consumers can now experience \360 content in mobile phones, desktops and, more importantly, the new arising medium -- Virtual Reality (VR) -- headsets.

Depth and/or geometry extraction from omnidirectional content has been approached similar to traditional 2D content via omnidirectional stereo \cite{li2008binocular,ma20153d,pathak2016dense,li2005spherical} and SfM\cite{huang20176} analytical solutions. There are inherent problems though to applying learning based methods to \360 content as a result of its acquisition process that inhibits the creation of high quality datasets. Coupling them with \360 LIDARs would produce low resolution depths and would also insert the depth sensor into the content itself, a drawback that also exists when aiming to acquire stereo datasets. One alternative would be to manually re-position the camera but that would be tedious and error prone as a consistent baseline would not be possible.

In this work, we train a CNN to learn to estimate a scene's depth given an omnidirectional (equirectangular) image as input\footnote{We use the terms omnidirectional image, \360 image, spherical panorama and equirectangular image interchangeably in this document.}. To circumvent the lack of available training data we resort to re-using existing 3D datasets and re-purposing them for use within a \360 context. This is accomplished by generating diverse \360 views via rendering. We use this dataset for learning to infer depth from omnidirectional content. In summary, our contributions are the following:

\begin{enumerate}
\item We present the first, to the authors' knowledge, learning based dense depth estimation method that was trained with and operates directly on omnidirectional content in the form of equirectangular images.
\item We offer a dataset consisting of \360 color images paired with ground truth \360 depth maps in equirectangular format. The dataset is available online\footnote{\label{foot:dataset}\url{\datasetURL}}.
\item We propose and validate, a CNN auto-encoder architecture specifically designed for estimating depth directly on equirectangular images.
\item We show how monocular depth estimation methods trained on traditional 2D images fall short or produce low quality results when applied to equirectangular inputs, highlighting the need for learning directly on the \360 domain.
\end{enumerate}

\section{Related Work}
\label{sec:related}
Since this work aims to learn the task of omnidirectional dense depth estimation, and given that - to the authors' knowledge - no other similar work exists, we first review non-learning based methods for geometric scene understanding based on \360 images. We then examine learning based approaches for spherical content and, finally, present recent monocular dense depth estimation methods.

\subsection{Geometric understanding on \360 images}
\label{sec:360geometry}
Similar to pinhole projection model cameras, the same multi-view geometry \cite{hartley2000multiple} principles apply to \360 images. By observing the scene from multiple viewpoints and establishing correspondences between them, the underlying geometrical structure can be estimated. For \360 cameras the conventional binocular or multi-view stereo \cite{furukawa2015multi} problem is reformulated to binocular or multi-view spherical stereo \cite{li2008binocular} respectively, by taking into account the different projection model and after defining the disparity as angular displacements. By estimating the disparity (i.e. depth), complete scenes can be 3D reconstructed from multiple \cite{kim2013scenereconstruction,li2005spherical} or even just two \cite{ma20153d,pathak2016dense} spherical viewpoints. However, all these approaches require multiple \360 images to estimate the scene's geometry. Recently it was shown that \360 videos acquired with a moving camera can be used to 3D reconstruct a scene's geometry via SfM \cite{huang20176} and enable 6 DOF viewing in VR headsets. 

There are also approaches that require only a single image to understand indoors scenes and estimate their layout. PanoContext \cite{zhang2014panocontext}, generates a 3D room layout hypothesis given an indoor \360 image in equirectangular format. With the estimations being bounding boxes, the inferred geometry is only a coarse approximation of the scene. Similar in spirit, the work of Yang et al. \cite{yang2016efficient} generates complete room layouts from panoramic indoor images by combining superpixel information, vanishing points estimation and a geometric context prior under a Manhattan world assumption. However, focusing on room layout estimation, it is unable to recover finer details and structures of the scene. Another similar approach \cite{xu2017pano2cad} addresses the problem of geometric scene understanding from another perspective. Under a maximum a posteriori estimation it unifies semantic, pose and location cues to generate CAD models of the observed scenes. Finally, in \cite{kim2016room} a spherical stereo pair is used to estimate  both the room layout but also object and material attributes. After retrieving the scene's depth by stereo matching and subsequently calculating the normals, the equirectangular image is projected to the faces of a cube that are then fed to a CNN whose \textbf{object} predictions are fused into the \360 image to finally reconstruct the 3D layout. 

\subsection{Learning for \360 images}
\label{sec:360learning}
One of the first approaches to estimate distances purely from omnidirectional input \cite{plagemann2010nonparametric} under a machine learning setting utilized Gaussian processes. Instead of estimating the distance of each pixel, a range value per image column was predicted to  drive robotic navigation. Nowadays, with the establishment of CNNs, there are two straightforward ways to apply current CNN processing pipelines to spherical input. Either directly on a projected (typically equirectangular) image, or by projecting the spherical content to the faces of a cube (cubemap) and running the CNN predictions on them, which are then merged by back-projecting them to the spherical domain. The latter approach was selected by an artistic style transfer work \cite{ruder2017artistic}, where each face was re-styled separately and then the cubemap was re-mapped back to the equirectangular domain. Likewise, in SalNet360 \cite{monroy2017salnet360}, saliency predictions on the cube's faces are refined using their spherical coordinates and then merged back to \360. The former approach, applying a CNN directly to the equirectangular image, was opted for in \cite{zhang2017learning} to increase the dynamic range of outdoor panoramas.

More recently, new techniques for applying CNNs to omnidirectional input were presented. Given the difficulty to model the projection's distortion directly in typical CNNs as well as achieve invariance to the viewpoint's rotation, the alternative pursued by \cite{frossard2017graph} is based on graph-based deep learning. Specifically they model distortion directly into the graph's structure and apply it to a classification task. A novel approach taken in \cite{su2017learning} is to learn appropriate convolution weights for equirectangular projected spherical images by transferring them from an existing network trained on traditional 2D images. This conversion from the 2D to the \360 domain is accomplished by enforcing consistency between the predictions of the 2D projected views and those in the \360 image. Moreover, recent work on convolutions \cite{jeon2017active,dai2017deformable} that in addition to learning their weights also learn their shape, are very well suited for learning the distortion model of spherical images, even though they have only been applied to fisheye lenses up to now \cite{deng2018restricted}. Finally, very recently, Spherical CNNs were proposed in \cite{cohen2017convolutional,cohen2018spherical} that are based in a rotation-equivariant definition of spherical cross-correlation. However these were only demonstrated in classification and single variable regression problems. In addition, they are also applied in the spectral domain while we formulate our network design for the spatial image domain.
   
\subsection{Monocular depth estimation}
\label{sec:monocular}
Depth estimation from monocular input has attracted lots of interest lately. While there are some impressive non learning based approaches \cite{ranftl2016dense,liu2014discrete,karsch2016depth}, they come with their limitations, namely reliance on optical flow and relevance of the training dataset. Still, most recent research has focused on machine learning to address the ill-posed depth estimation problem. Initially, the work of Eigen et al. \cite{eigen2014depth} trained a CNN in a coarse-to-fine scheme using direct depth supervision from RGB-D images. In a subsequent continuation of their work \cite{eigen2015predicting}, they trained a multi-task network that among predicting semantic labels and normals, also estimated a scene's depth. Their results showed that jointly learning the tasks achieved higher performance due to their complementarity. In a recent similar work \cite{ren-cvpr2018}, a multi-task network that among other modalities also estimated depth, was trained using synthetic data and a domain adaptation loss based on adversarial learning, to increase its robustness when running on real scenes. Laina et al. \cite{laina2016deeper} designed a directly supervised fully convolutional residual network (FCRN) with  novel up-projection blocks that achieved impressive results for indoor scenes and was also used in a SLAM pipeline \cite{tateno2017cnnslam}. 

Another body of work focused on applying Conditional Random Fields (CRFs) to the depth estimation problem. Initially, the output of a deep network was refined using a hierarchical CRF \cite{li2015depth}, with Liu et al. \cite{liu2016learning} further exploring the interplay between CNNs and CRFs for depth estimation in their work. Recently, multi-scale CRFs were used and trained in an end-to-end manner along with the CNN \cite{xu2017multi}. Dense depth estimation has also been addressed as a classification problem. Since perfect regression is usually impossible, dense probabilities were estimated in \cite{chakrabarti2016depth} and then optimized to estimate the final depth map. Similarly, in \cite{li2017monocular} and \cite{cao2017estimating} depth values were discretized in bins and densely classified, to be afterwards refined either via a hierarchical fusion scheme or through the use of a CRF respectively. Taking a step further, a regression-classification cascaded network was proposed in \cite{fu2017compromise} where a low spatial resolution depth map was regressed and then refined by a classification branch.

The concurrent works of Garg et al. \cite{garg2016unsupervised} and Godard et al. \cite{monodepth17} showed that unsupervised learning of the depth estimation task is possible. This is accomplished by an intermediate task, view synthesis, and allowed training by only using stereo pair input with known baselines. In a similar fashion, using view synthesis as the main supervisory signal, learning to estimate depth was  also achieved by training with pure video sequences in a completely unsupervised manner \cite{zhou2017unsupervised,wang2017learning,mahjourian2018unsupervised,yang2017unsupervised,yin2018geonet}. Another novel unsupervised depth estimation method relies on aperture supervision \cite{srinivasan2017aperture} by simply acquiring training data in various focus levels. Finally, in \cite{chen2016single} it was shown that a CNN can be trained to estimate depth from monocular input with only relative depth annotations.

\section{Synthesizing Data}
\label{sec:datagen}
End-to-end training of deep networks requires a large amount of annotated ground truth data. While for typical pinhole camera datasets this was partly addressed by using depth sensors \cite{silberman2012indoor} or laser scanners \cite{saxena2009make3d} such an approach is impractical for spherical images due to a larger diversity in resolution for \360 cameras and laser scanners, and because each \360 sensor would be visible from the other one. As much as approaches like the one employed in \cite{matzen2017low} could be used to in-paint the sensor regions, these would still be the result of an algorithmic process and not the acquisition process itself, potentially introducing errors and artifacts that would reduce the quality of the data. This also applies to unsupervised stereo approaches that require the simultaneous capture of the scene from two viewpoints. Although one could re-position the same sensor to acquire clean panoramas, a consistent baseline would not be possible. More recently, unsupervised approaches for inferring a scene's depth  have emerged that are trained with video sequences. However, they assume a moving camera as they rely on view synthesis as the supervisory signal which is not a typical setting for indoors \360 captures, but for action camera like recordings.

\DATASET Instead, we rely on generating a dataset with ground truth depth by synthesizing both the color and the depth image via rendering. To accomplish that we leverage the latest efforts in creating publicly available textured 3D datasets of indoors scenes. Specifically, we use two computer generated (CG) datasets, SunCG \cite{song2016ssc} and SceneNet \cite{handa2016scenenet}, and two realistic ones acquired by scanning indoor buildings, Stanford2D3D \cite{armeni2017joint,armeni20163d} and Matterport3D \cite{Matterport3D}. We use a path tracing renderer\footnote{\url{https://www.cycles-renderer.org}} to render our dataset by placing a spherical camera and a uniform point light at the same position $\mathbf{c} \in \mathbb{R}^3$ in the scene. We then acquire the rendered image $I(\mathbf{p}) \in \mathbb{R}, \mathbf{p} = (u, v) \in \mathbb{N}^2$, as well as the underlying $z$-buffer that was generated as a result of the graphics rendering process, that serves as the ground truth depth $D(\mathbf{p}) \in \mathbb{R}$. It should be noted that unlike pinhole camera model images, the $z$-buffer in this case does not contain the $z$ coordinate value of the 3D point $\mathbf{v}(\mathbf{p}) \in \mathbb{R}^3$, corresponding to pixel $\mathbf{p}$, but instead the 3D point's radius $r = \| \mathbf{v} - \mathbf{c} \|$, in the camera's spherical coordinate system. 

For the two CG indoors datasets we place the camera and the light at the center of each house, while for the two scanned indoors datasets we use the pose information available (estimated during the scanning process) and thus, for each building we generate multiple \360 data samples. Given that the latter two datasets were scanned, their geometries contain holes or inaccurate/coarse estimations, and also have lighting information baked into the models. On the other hand, the CG datasets contain perfect per pixel depth but lack the realism of the scanned datasets, creating a complementary mix. However, as no scanning poses are available, the centered poses may sometimes be placed within or on top of objects and we also observed missing information in some scenes (e.g. walls/ceilings) that, given SunCG's size, are impractical to manually correct.

For each pose, we augment the dataset by rotating the camera in \degree{90} resulting in 4 distinct viewpoints per pose sample. Given the size of SunCG, we only utilize a subset of it and end up using $\textbf{11118}$ houses, resulting in a $24.36\%$ utilization. The remaining three datasets are completely rendered. This results in a total of $\textbf{94098}$ renders and $\textbf{23524}$ unique viewpoints. Our generated \textit{\Dataset}dataset contains a mix of synthetic and realistic \360 color $I$ and depth $D$ image data in a variety of indoors contexts (houses, offices, educational spaces, different room layouts) and is publicly available at \url{\datasetURL}.

\begin{figure}[!t]
    \centering
    \begin{rotate}{90}{\scriptsize Matterport3D}\end{rotate}
    \begin{subfigure}[t]{0.3\textwidth}
        \centering
    	\includegraphics[scale = 0.085]{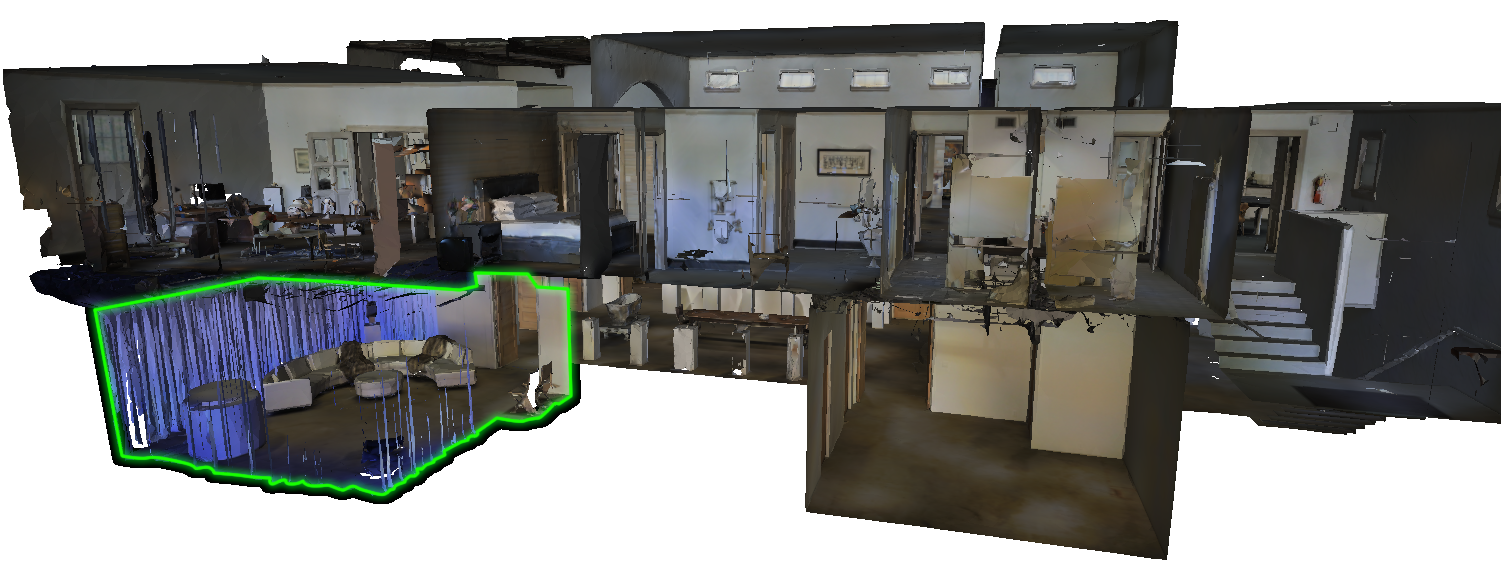}
    \end{subfigure}
	\begin{subfigure}[t]{0.22\textwidth}
	    \centering
    	\includegraphics[scale = 0.15]{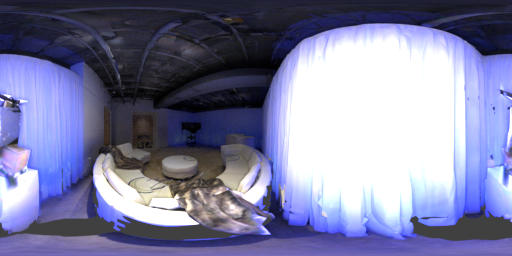}
    \end{subfigure}
	\begin{subfigure}[t]{0.22\textwidth}
   	    \centering
    	\includegraphics[scale = 0.15]{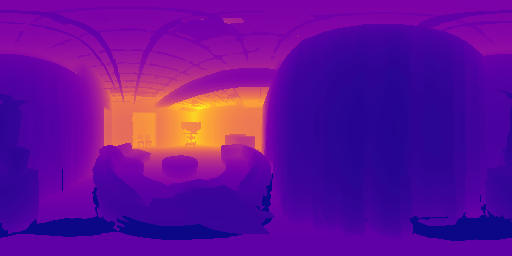}
    \end{subfigure}
    \begin{subfigure}[t]{0.22\textwidth}
        \centering
    	{
    		\setlength{\fboxsep}{0pt}%
			\setlength{\fboxrule}{0.1pt}%
			\fbox{\includegraphics[scale = 0.15]{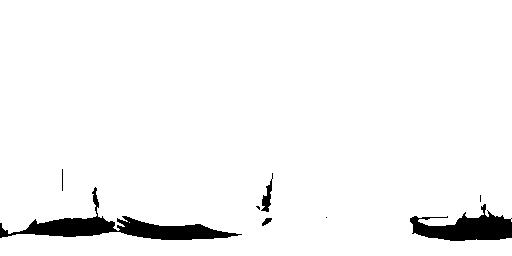}}
    	}
    \end{subfigure}
    
    \begin{rotate}{90}{\scriptsize Stanford2D3D}\end{rotate}
    \begin{subfigure}[t]{0.3\textwidth}
        \centering
    	\includegraphics[scale = 0.08]{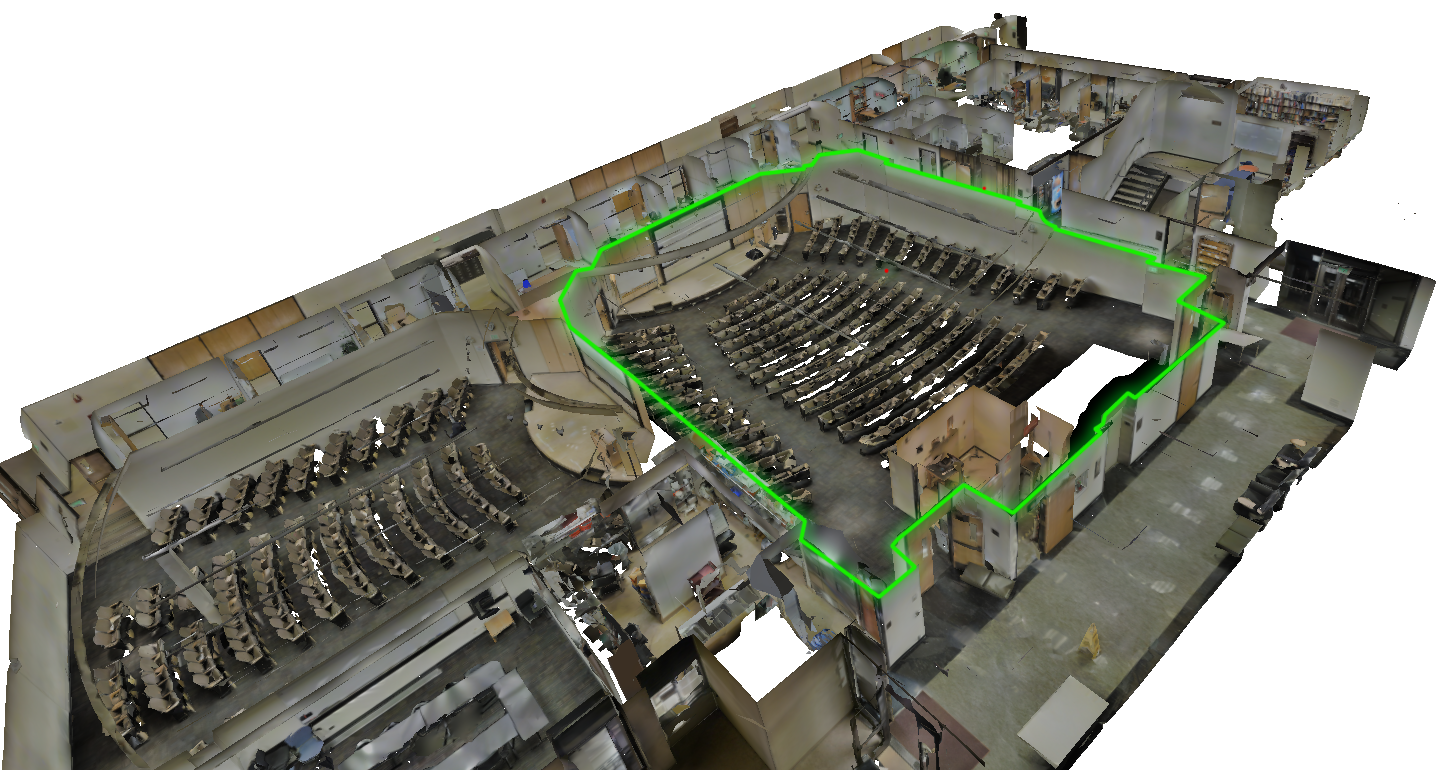}
    \end{subfigure}
	\begin{subfigure}[t]{0.22\textwidth}
	    \centering
    	\includegraphics[scale = 0.15]{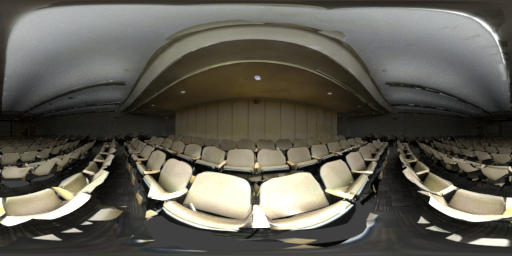}
    \end{subfigure}
	\begin{subfigure}[t]{0.22\textwidth}
	    \centering
    	\includegraphics[scale = 0.15]{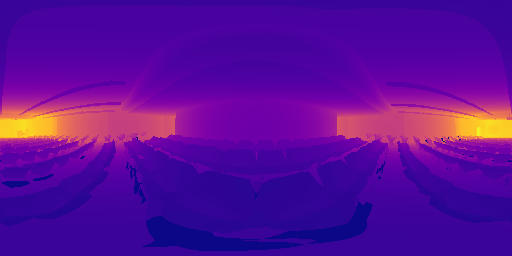}
    \end{subfigure}
    \begin{subfigure}[t]{0.22\textwidth}
        \centering
    	{
    		\setlength{\fboxsep}{0pt}%
			\setlength{\fboxrule}{0.1pt}%
			\fbox{\includegraphics[scale = 0.15]{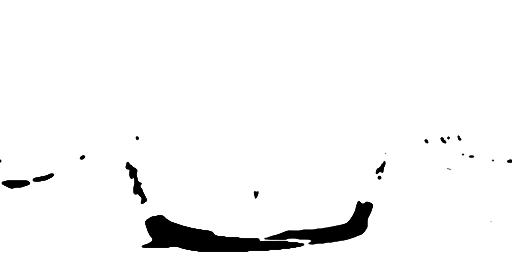}}
    	}
    \end{subfigure}
    
    \begin{rotate}{90}{\scriptsize SunCG}\end{rotate}
    \begin{subfigure}[t]{0.3\textwidth}
        \centering
    	\includegraphics[scale = 0.085]{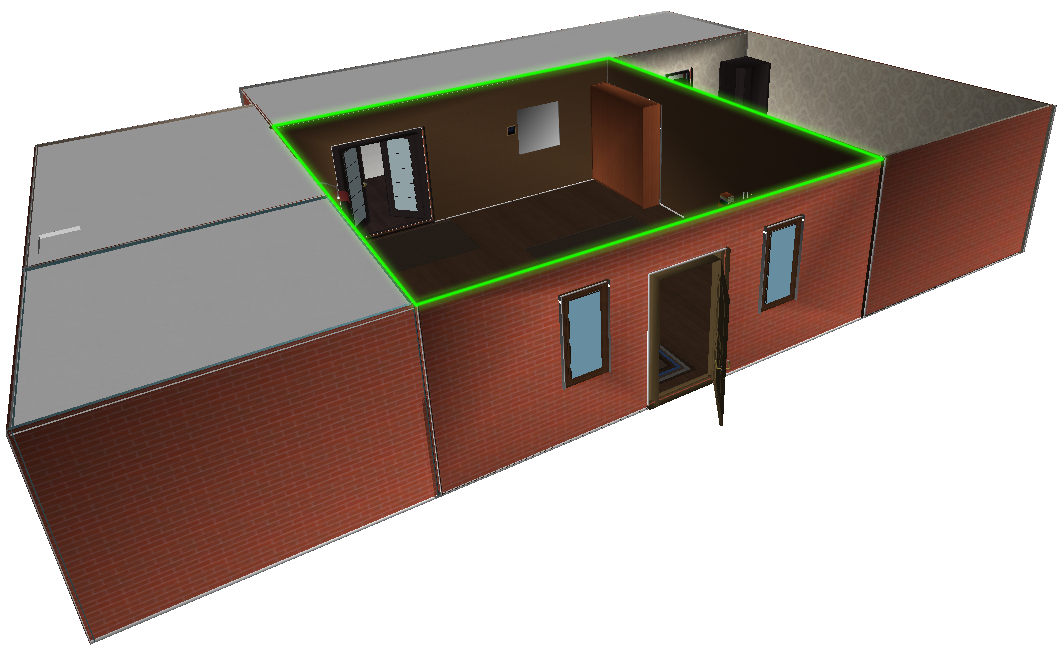}
    \end{subfigure}
	\begin{subfigure}[t]{0.22\textwidth}
	    \centering
    	\includegraphics[scale = 0.15]{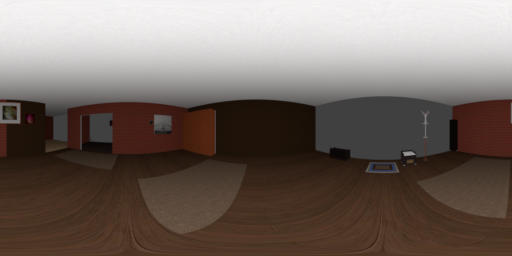}
    \end{subfigure}
	\begin{subfigure}[t]{0.22\textwidth}
	    \centering
    	\includegraphics[scale = 0.15]{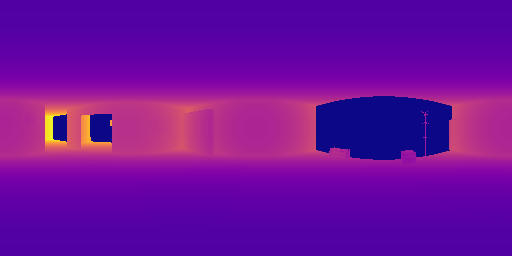}
    \end{subfigure}
    \begin{subfigure}[t]{0.22\textwidth}
        \centering
    	{
    		\setlength{\fboxsep}{0pt}%
			\setlength{\fboxrule}{0.1pt}%
			\fbox{\includegraphics[scale = 0.15]{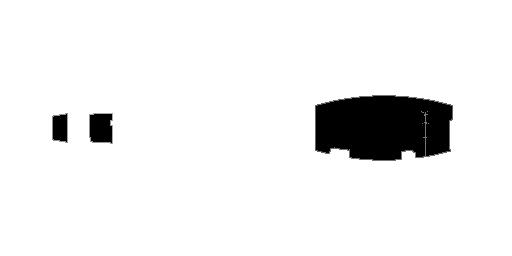}}
    	}
    \end{subfigure}
    \caption{Example renders from our dataset, from left to right: the 3D building with a green highlight denoting the rendered scene, color output, corresponding depth map, and the binary mask depicting the missing regions in black.}
    \label{fig:datagen}
\end{figure}

\section{Omnidirectional Depth Estimation}
\label{sec:omnidepth}
The majority of recent CNN architectures for dense estimation follow the autoencoder structure, in which an encoder encodes the input, by progressively decreasing its spatial dimensions, to a representation of much smaller size, and a decoder that regresses to the desired output by upscaling this representation.

We use two encoder-decoder network architectures that are structured differently. The first resembles those found in similar works in the literature \cite{monodepth17,laina2016deeper}, while the second is designed from scratch to be more suitable for learning with \360 images. Both networks are fully convolutional \cite{long2015fully} and predict an equirectangular depth map with the only input being a \360 color image in equirectangular format. We use ELUs \cite{clevert2015fast} as the activation function which also remove the need for batch normalization \cite{ioffe2015batch} and its added computational complexity. 

\begin{figure}[!htbp]
	\centering
	\includegraphics[scale = 0.33]{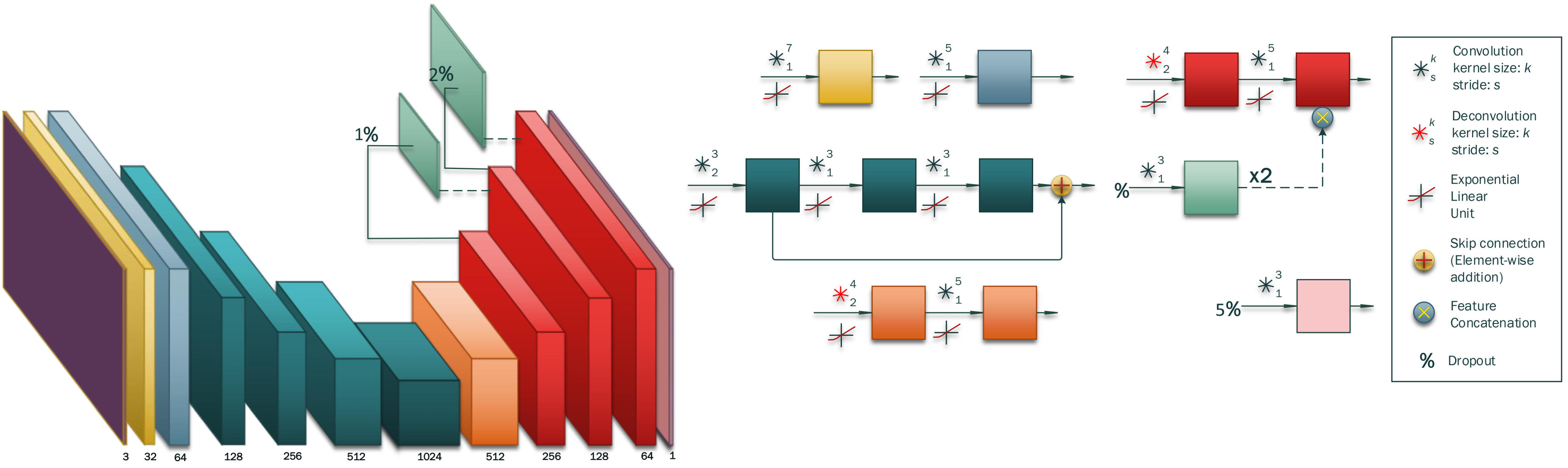}
    \caption{\textbf{\unbal Architecture: }The encoder consists of two input preprocessing blocks, and four down-scaling blocks (dark green). The former are single convolutional (conv) layers while the latter consist of a strided conv and two more regular convs with a skip/residual connection. The decoder contains one upscaling block (orange) and three up-prediction blocks (red), followed by the prediction layer (pink). Up-scaling is achieved with a strided deconv followed by a conv, and similarly, up-predictions additionally branch out to estimate a depth prediction at the corresponding scale with an exta conv that is concatenated with the features of the next block's last layer.}
    \label{fig:unbalResnet}
\end{figure}

\UNBAL In this unbalanced ResNet, the encoding and decoding parts are not symmetrical, with the decoder being shallower. The encoder is built with skip connections \cite{he2016deep}, a technique that helps when training deeper architectures by preventing gradient degradation, allowing for larger receptive fields. More detailed architectural information is presented in Fig. \ref{fig:unbalResnet} where the network is decomposed into processing blocks.

\RECTDIL Omnidirectional images differ from traditional images in the sense that they capture global (full \360) visual information and, when in equirectangular format, suffer from high distortions along their $y$ (i.e. latitude) axis. Therefore, the second architecture's design aims to exploit and address these properties of spherical panoramas while keeping some of the desirable properties of \unbal like skip connections. Capturing the \360 image's global context is achieved by increasing the effective receptive field (RF) of each neuron by utilizing dilated convolutions \cite{yu2017dilated}. Instead of progressive downscaling as in most depth estimation networks and similarly \unbalnospace, we only drop the spatial dimensions by a factor of 4. Then, inspired by \cite{van2018light}, we use progressively increasing dilations to increase the RF to about half the input's spatial dimensions and increase global scene understanding. In addition, within each dilation block we utilize \conv11 convolutions to reduce the spatial correlations of the feature maps.

\begin{figure}[!htbp]
	\centering
	\includegraphics[scale=0.33]{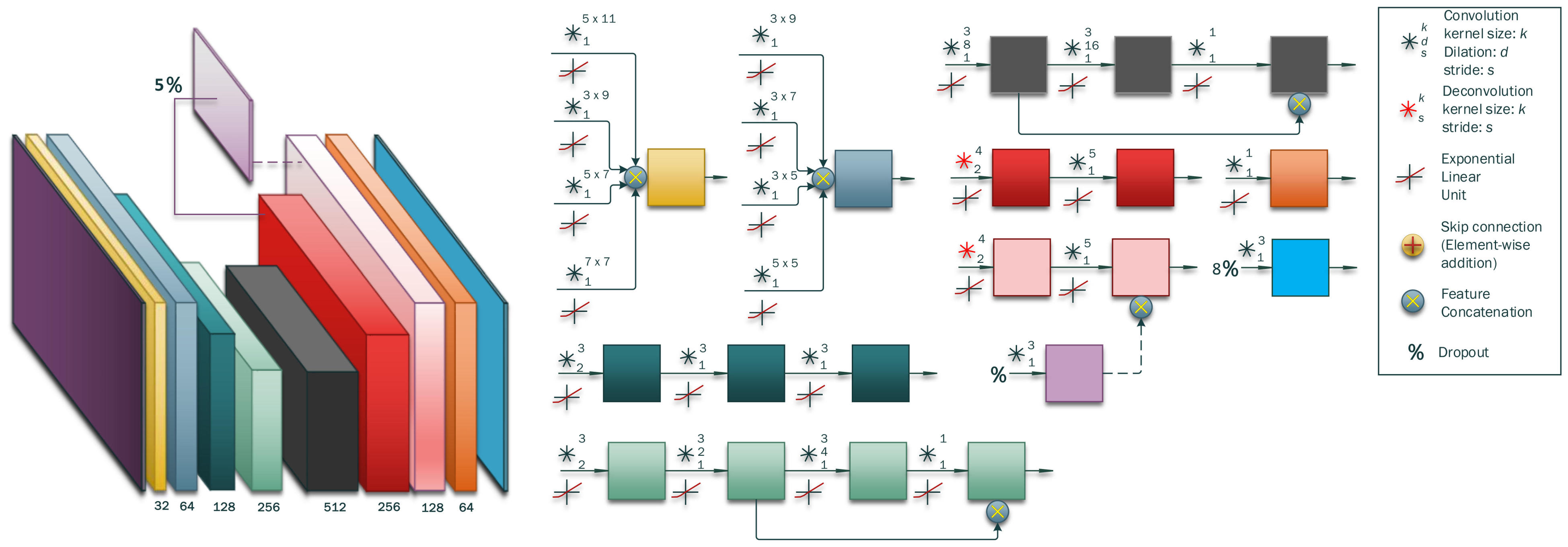}
    \caption{\textbf{\rectdil Architecture: }The encoder consists of two preprocessing blocks (yellow and blue) and a downscaling block (dark green), followed by two increasing dilation blocks (light green and black). The preprocessing blocks concatenate features produced by convolutions (convs) with different filter sizes, accounting for the equirectangular projection's varying distortion factor. The down-scaling block comprises a strided and two regular convs. }
    \label{fig:rectDilnet}
\end{figure}

The distortion factor of spherical panoramas increases towards the sphere's poles and is therefore different for every image row. This means that information is scattered horizontally, as we vertically approach the two poles. In order to account for this varying distortion we alter our input blocks, as their features are closer to natural image ones (e.g. edges). Following \cite{su2017learning}, where 2D CNN filters are transfered into distorted (practically rectangular) row-wise versions to increase performance when applied to the \360 domain, we use rectangle filters along with traditional square filters and vary the resolution of the rectangle ones to account for different distortion levels. However, this variation is done while also preserving the area of the filter to be as close as possible to the original square filter's. The outputs of the rectangle and square filters are concatenated while preserving the overall output feature count. The detailed architecture is presented in Fig. \ref{fig:rectDilnet}.

\textbf{Training Loss: } Given that we synthesize perfect ground truth depth annotations $D_{gt}$, as presented in Section \ref{sec:datagen}, we take a completely supervised approach. Even though most approaches using synthetic data fail to generalize to realistic input, our dataset contains an interesting mix of synthetic (CAD) renders as well as realistic ones. The scanned data are acquired from real environments and, as a result, their renders are very realistic. Following previous work, we predict depth $D_{pred}^{s}$ against downscaled versions of the ground truth data $D_{gt}^{s}$ at multiple scales (with $s$ being the downscaling factor) and upsample these predictions using nearest neighbor interpolation to later concatenate them with the subsequent higher spatial dimension feature maps. We also use the dropout technique \cite{srivastava2014dropout} in those layers used to produce each prediction. Further, we use L2 loss for regressing the dense depth output $E_{depth}(\mathbf{p}) = \| D_{gt}(\mathbf{p}) - D_{pred}(\mathbf{p}) \|^2$ and additionally add a smoothness term $E_{smooth}(\mathbf{p}) = \| \nabla D(\mathbf{p}) \|^2$ for the predicted depth map by minimizing its gradient. 

Although our rendered depth maps are accurate in terms of depth, in practice there are missing regions in the rendered output. These are either because of missing information in the CAD models (e.g. walls/ceilings) or the imperfect process of large scale 3D scanning, with visual examples illustrated in Fig. \ref{fig:datagen}. These missing regions/holes manifest as a specific color ("clear color"), selected during rendering, in the rendered image and as infinity ("far") values in the rendered depth map. As these outlier values will destabilize the training process, we ignore them during backpropagation by using a per pixel $\textbf{p}$ binary mask $M(\textbf{p})$ that is zero in these missing regions. This allows us to train the network even with incomplete or slightly inaccurate/erroneous 3D models. Thus, our final loss function is:
\begin{equation}
	E_{loss}(\mathbf{p}) = \sum_s{\alpha_s M(\mathbf{p}) E_{depth}(\mathbf{p})} + \sum_s{\beta_s M(\mathbf{p}) E_{smooth}(\mathbf{p})} \, ,
\end{equation}
where $\alpha_s, \beta_s$ are the weights for each scale of the depth and smoothing term.

\section{Results}
\label{sec:results}
We evaluate the performance of our two \360 depth estimation networks by first conducting an intra assessment of the two models and then offering quantitative comparisons with other depth estimation methods. Finally, we present comparative qualitative results in unseen, realistic data of everyday scenes.

\begin{table}[!t]
\centering
\caption{Quantitative results of our networks for \360 dense depth estimation.}
\label{tab:resOurs}
\resizebox{\textwidth}{!}{
\begin{tabular}{l|c|cccc|ccc}
\hline
Network & Tested on & Abs Rel \(\downarrow\) & Sq Rel  \(\downarrow\)& RMS    \(\downarrow\)& RMSlog \(\downarrow\)& \(\delta < 1.25\) \(\uparrow\)& \(\delta < 1.2^2\) \(\uparrow\)& \(\delta < 1.25^3\) \(\uparrow\)\\
\hline \hline

\unbal & Test set & 0.0835 & 0.0416 & 0.3374 & 0.1204 & 0.9319 & 0.9889 & 0.9968 \\

\rectdil & Test set & \textbf{0.0702} & \textbf{0.0297} & \textbf{0.2911} & \textbf{0.1017} & \textbf{0.9574} & \textbf{0.9933} & \textbf{0.9979} \\

\unbal & SceneNet & 0.1218 & 0.0727 & 0.4066 & 0.1538 & 0.8598 & 0.9815 & 0.9962 \\

\rectdil & SceneNet & 0.1077 & 0.699 & 0.3572 & 0.1386 & 0.8965 & 0.9879 & 0.9971     \\
\hline

\unbal-S2R & Stanford & 0.1226 & 0.0768 & 0.489 & 0.1667 & 0.8593 & 0.9756 & 0.9942 \\

\rectdil-S2R & Stanford & 0.0824 & 0.0457 & 0.3998 & 0.1229 & 0.928 & 0.9879 & 0.9971 \\

\unbal-S2R & SceneNet & 0.1448 & 0.0991 & 0.517 & 0.1792 & 0.7898 & 0.9761 & 0.9935 \\

\rectdil-S2R & SceneNet & 0.1079 & 0.0644 & 0.3778 & 0.1404 & 0.8966 & 0.9866 & 0.996 \\

\hline \hline
\end{tabular}
}
\end{table}

\textbf{Training Details}: Our networks are trained using Caffe \cite{jia2014caffe} on a single NVIDIA Titan X. We use Xavier weight initialization \cite{glorot10axavier} and ADAM \cite{kingma2014adam} as the optimizer with its default parameters $[\beta_1,\beta_2,\epsilon] = [0.9, 0.999, 10^-8]$ and an initial learning rate of $0.0002$. Our input dimensions are $512 \times 256$ and are given in equirectangular format, with our depth predictions being equal sized. 

We split our dataset into corresponding train and tests sets as follows: (i) Initially we remove 1 complete area from Stanford2D3D, 3 complete buildings from Matterport3D and 3 CAD scenes from SunCG for our test set totaling 1,298 samples. (ii) We skip SceneNet entirely and use it as our validation set. (iii) Then, from the remaining SunCG, Stanford2D3D and Matterport3D samples we automatically remove scenes which contain regions with very large or small depth values ($>5\%$ of total image area above $20m$ or under $0.5m$). Finally, we are left with a train-set that consists of 34,679\footnote{Only a subset of SunCG was used by prioritizing larger scenes given the length of the rendering process. However, a larger subset is publicly available.} RGB \360 images along with their corresponding ground truth depth map annotations. Our loss weights for \unbal are $[\alpha_1,\alpha_2,\alpha_4,\beta_1] = [0.445,0.275,0.13,0.15]$, and for \rectdil they are $[\alpha_1,\alpha_2,\beta_1,\beta_2] = [0.535,0.272,0.134,0.068]$. For quantitative evaluation we use the same error metrics as previous works \cite{eigen2014depth,eigen2015predicting,monodepth17,laina2016deeper,liu2016learning} (arrows next to each metric in the tables denote the direction of better performance).

\textbf{Model Performance: }Table \ref{tab:resOurs} presents the results of our two models in our test set, and in the unseen synthetic SceneNet generated data, after training for 10 epochs in all of our train set. We observe that \rectdil -- which was designed with \360 input in mind -- performs better than the standard \unbal even with far fewer parameters ($\sim8.8M$ vs $\sim51.2M$). In order to assess their efficacy and generalization capabilities we perform a leave-one-out evaluation. We train both networks initially only in the synthetic SunCG generated data for 10 epochs, and then finetune them in the realistic Matterport3D generated data for another 10 epochs. This train is suffixed with "-S2R". We then evaluate them in the entirety of the Stanford2D3D generated dataset, as well as in the SceneNet one. Comparable results to the previous train with all datasets are observed. Again, \rectdil outperforms \unbal -- albeit both perform slightly worse as expected due to being trained with less amount of data.

The increased performance of \rectdil against \unbal in every error metric or accuracy, can be attributed to its larger RF, which for \360 images is very important as it allows the network to capture the global context more efficiently\footnote{Varying RF experiments supporting this claim can be found in the supplement.}. Despite the fact that \unbal is much deeper than \rectdil and significantly drops the input's spatial dimensions, \rectdil still achieves a larger receptive field. Specifically, \unbal has a $190 \times 190$ RF compared to that of \rectdil which is $266 \times 276$. In addition, \rectdil drops the input's spatial dimensions only by a factor of 4, maintaining denser information in the extracted features. 

\begin{table}[!t]
	\caption{Quantitative results against other monocular depth estimation models.}
    \label{tab:resOthers}    
	\centering
    \resizebox{\textwidth}{!}{
    \begin{tabular}{ll  | c c c c | c  c c  }
    	\hline
   		&Network & Abs Rel\(\downarrow\) & Sq Rel \(\downarrow\)& RMS \(\downarrow\)& RMS(log) \(\downarrow\)& \( \delta < 1.25\) \(\uparrow\)& \( \delta < 1.25^2\) \(\uparrow\)& \( \delta < 1.25^3\) \(\uparrow\)\\
        \hline \hline
        &\unbal & 0.0835 & 0.0416 & 0.3374 & 0.1204 & 0.9319 & 0.9889 & 0.9968 \\
        
        &\rectdil &  \textbf{0.0702} & \textbf{0.0297} & \textbf{0.2911} & \textbf{0.1017} & \textbf{0.9574} & \textbf{0.9933} & \textbf{0.9979} \\
        
        \hline
        
        &Godard et al. \cite{monodepth17}  & 0.4747 & 2.3783 & 7.2097 & 0.82 & 0.297 & 0.79 & 0.751 \\
        
        &Laina et al. \cite{laina2016deeper} & 0.3181 & 0.4469 & 0.941 & 0.376 & 0.4922 & 0.7792 & 0.915 \\
        
        \begin{rotate}{90}{\tiny Equirect.}\end{rotate}&Liu et al. \cite{liu2016learning}& 0.4202 & 0.7597 & 1.1596 & 0.44 & 0.3889 & 0.7044 & 0.8774 \\
        
        \hline
        
        &Godard et al. \cite{monodepth17} & 0.2552 & 0.9864 & 4.4524 & 0.5087 & 0.3096 & 0.5506 & 0.7202 \\
        
        &Laina et al. \cite{laina2016deeper} & 0.1423 & 0.2544 & 0.7751 & 02497 & 0.5198 & 0.8032 & 0.9175 \\
        
        \begin{rotate}{90}{\tiny Cubemap}\end{rotate}&Liu et al. \cite{liu2016learning} & 0.1869 & 0.4076 & 0.9243 & 0.2961 & 0.424 & 0.7148 & 0.8705 \\
        
        \hline \hline
    \end{tabular}
   }
\end{table}
\begin{figure}[!t]
	\centering
    \begin{rotate}{90}{\tiny RGB}\end{rotate}
    \begin{subfigure}[t]{0.18\textwidth}
    	\includegraphics[scale=0.125]{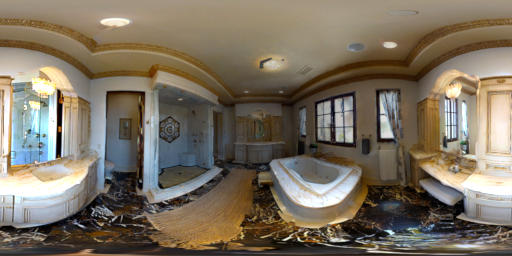}
    \end{subfigure}
    \begin{subfigure}[t]{0.18\textwidth}
    	\includegraphics[scale=0.125]{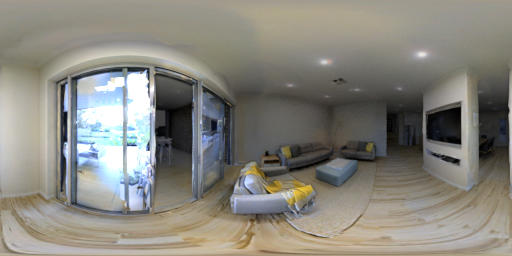}
    \end{subfigure}
    \begin{subfigure}[t]{0.18\textwidth}
    	\includegraphics[scale=0.125]{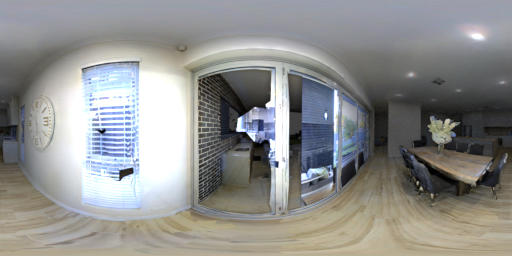}
    \end{subfigure}
    \begin{subfigure}[t]{0.18\textwidth}
    	\includegraphics[scale=0.125]{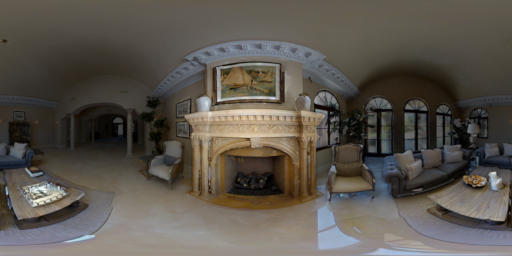}
    \end{subfigure}
    \begin{subfigure}[t]{0.18\textwidth}
    	\includegraphics[scale=0.125]{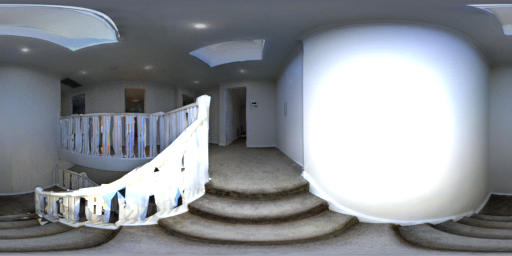}
    \end{subfigure}

    \begin{rotate}{90}{\scalebox{0.5}{Ground Truth}}\end{rotate}
    \begin{subfigure}[t]{0.18\textwidth}
    	\includegraphics[scale=0.125]{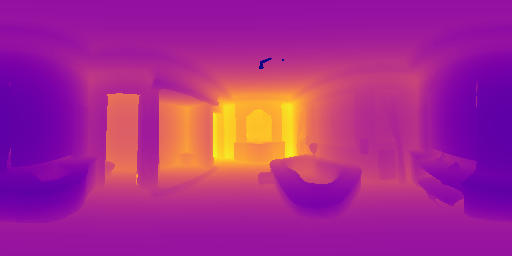}
    \end{subfigure}
    \begin{subfigure}[t]{0.18\textwidth}
    	\includegraphics[scale=0.125]{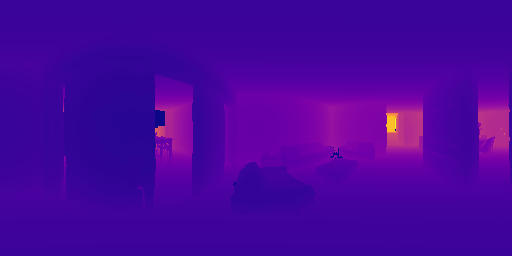}
    \end{subfigure}
    \begin{subfigure}[t]{0.18\textwidth}
    	\includegraphics[scale=0.125]{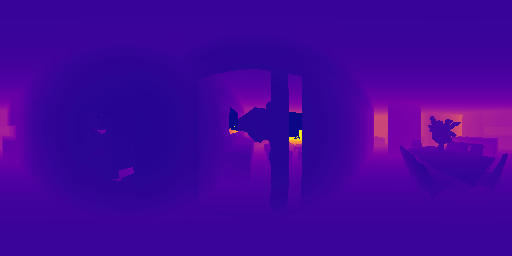}
    \end{subfigure}
    \begin{subfigure}[t]{0.18\textwidth}
    	\includegraphics[scale=0.125]{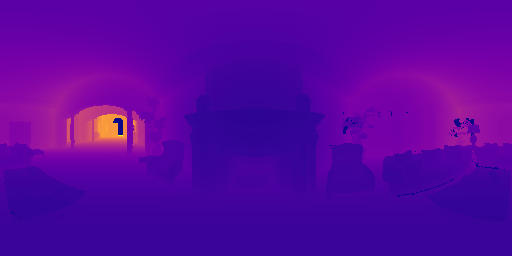}
    \end{subfigure}
    \begin{subfigure}[t]{0.18\textwidth}
    	\includegraphics[scale=0.125]{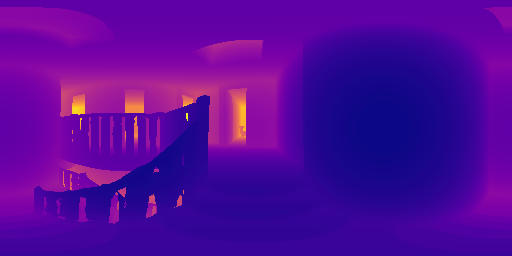}
    \end{subfigure}

    \begin{rotate}{90}{\scalebox{0.45}{Godard et al. \cite{monodepth17}}}\end{rotate}
    \begin{subfigure}[t]{0.18\textwidth}
    	\includegraphics[scale=0.125]{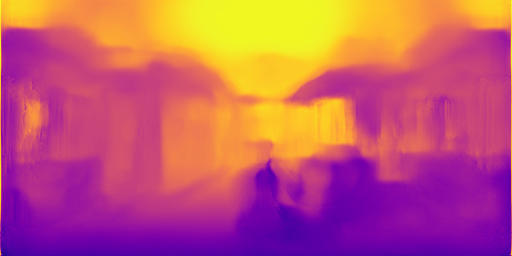}
    \end{subfigure}
    \begin{subfigure}[t]{0.18\textwidth}
    	\includegraphics[scale=0.125]{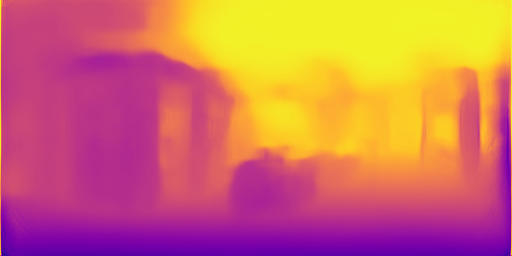}
    \end{subfigure}
    \begin{subfigure}[t]{0.18\textwidth}
    	\includegraphics[scale=0.125]{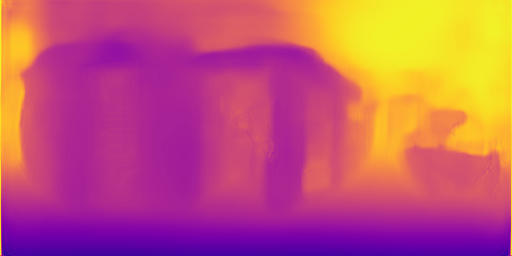}
    \end{subfigure}
    \begin{subfigure}[t]{0.18\textwidth}
    	\includegraphics[scale=0.125]{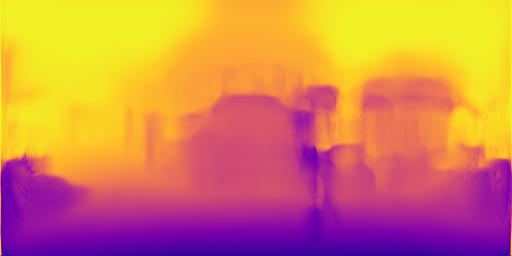}
    \end{subfigure}
    \begin{subfigure}[t]{0.18\textwidth}
    	\includegraphics[scale=0.125]{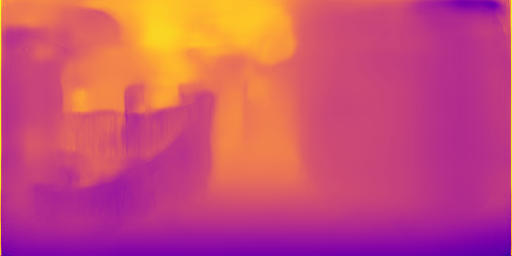}
    \end{subfigure}

    \begin{rotate}{90}{\scalebox{0.45}{Laina et al. \cite{laina2016deeper}}}\end{rotate}
    \begin{subfigure}[t]{0.18\textwidth}
    	\includegraphics[scale=0.125]{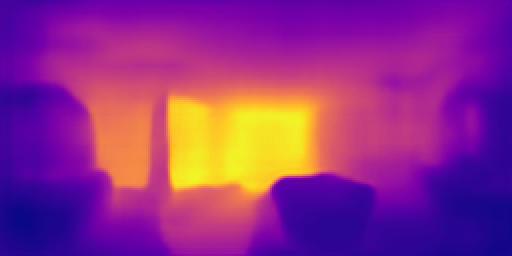}
    \end{subfigure}
    \begin{subfigure}[t]{0.18\textwidth}
    	\includegraphics[scale=0.125]{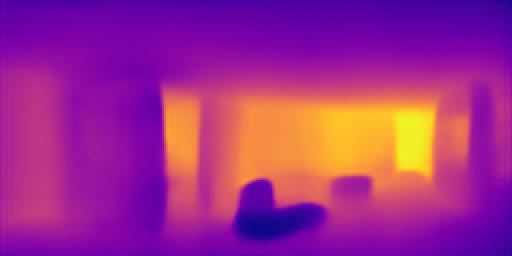}
    \end{subfigure}
    \begin{subfigure}[t]{0.18\textwidth}
    	\includegraphics[scale=0.125]{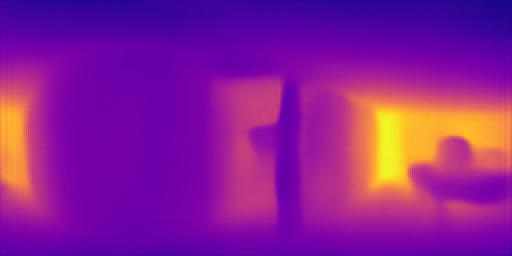}
    \end{subfigure}
    \begin{subfigure}[t]{0.18\textwidth}
    	\includegraphics[scale=0.125]{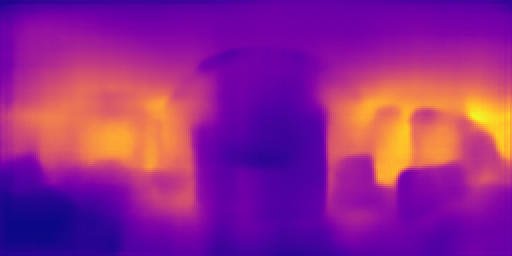}
    \end{subfigure}
    \begin{subfigure}[t]{0.18\textwidth}
    	\includegraphics[scale=0.125]{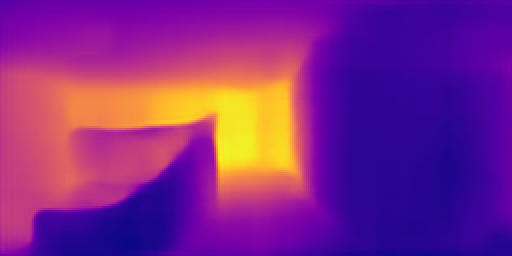}
    \end{subfigure}

    \begin{rotate}{90}{\scalebox{0.45}{Liu et al. \cite{liu2016learning}}}\end{rotate}
    \begin{subfigure}[t]{0.18\textwidth}
    	\includegraphics[scale=0.125]{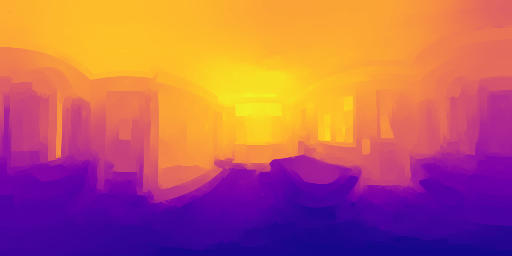}
    \end{subfigure}
    \begin{subfigure}[t]{0.18\textwidth}
    	\includegraphics[scale=0.125]{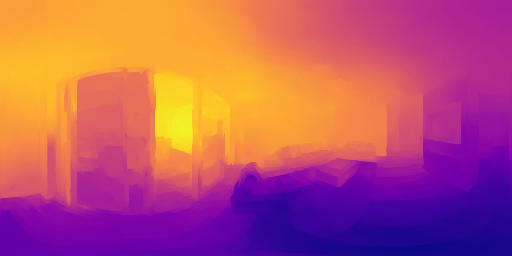}
    \end{subfigure}
    \begin{subfigure}[t]{0.18\textwidth}
    	\includegraphics[scale=0.125]{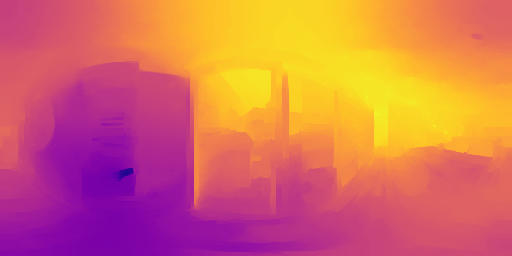}
    \end{subfigure}
    \begin{subfigure}[t]{0.18\textwidth}
    	\includegraphics[scale=0.125]{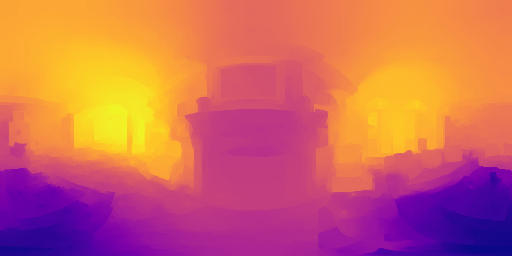}
    \end{subfigure}
    \begin{subfigure}[t]{0.18\textwidth}
    	\includegraphics[scale=0.125]{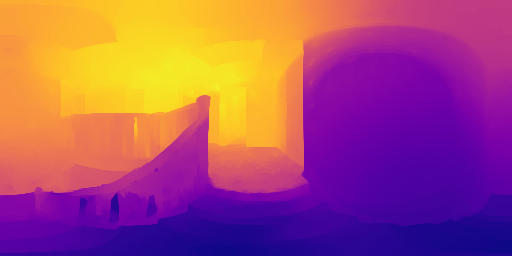}
    \end{subfigure}

    \begin{rotate}{90}{\scalebox{0.45}{Ours \unbal}}\end{rotate}
    \begin{subfigure}[t]{0.18\textwidth}
    	\includegraphics[scale=0.125]{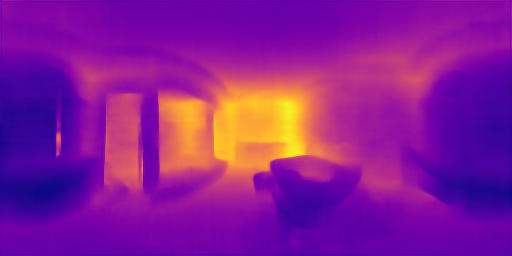}
    \end{subfigure}
    \begin{subfigure}[t]{0.18\textwidth}
    	\includegraphics[scale=0.125]{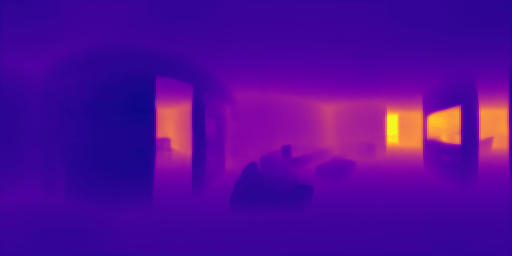}
    \end{subfigure}
    \begin{subfigure}[t]{0.18\textwidth}
    	\includegraphics[scale=0.125]{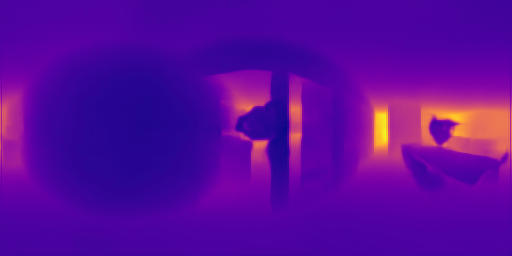}
    \end{subfigure}
    \begin{subfigure}[t]{0.18\textwidth}
    	\includegraphics[scale=0.125]{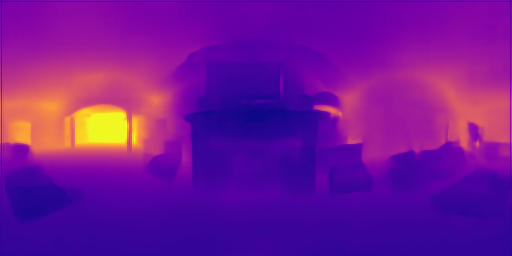}
    \end{subfigure}
    \begin{subfigure}[t]{0.18\textwidth}
    	\includegraphics[scale=0.125]{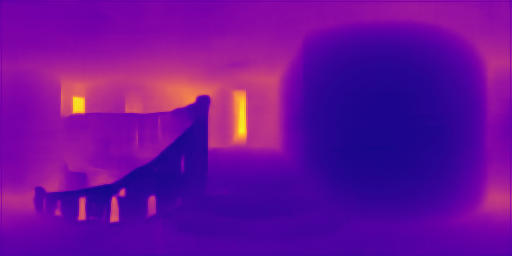}
    \end{subfigure}

    \begin{rotate}{90}{\scalebox{0.45}{Ours \rectdil}}\end{rotate}
    \begin{subfigure}[t]{0.18\textwidth}
    	\includegraphics[scale=0.125]{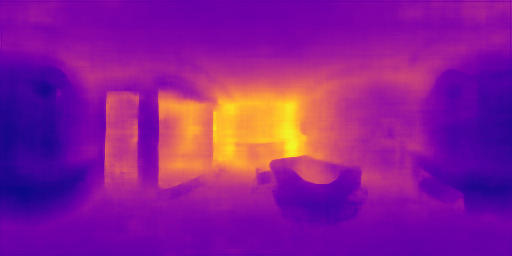}
    \end{subfigure}
    \begin{subfigure}[t]{0.18\textwidth}
    	\includegraphics[scale=0.125]{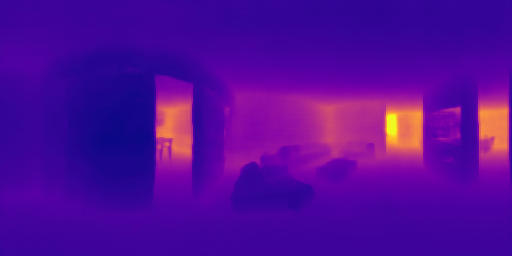}
    \end{subfigure}
    \begin{subfigure}[t]{0.18\textwidth}
    	\includegraphics[scale=0.125]{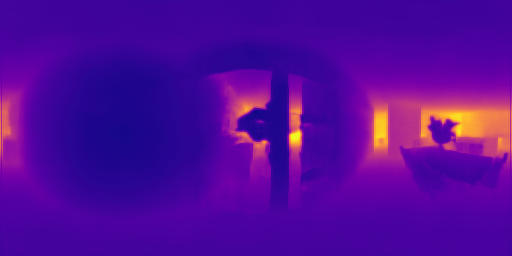}
    \end{subfigure}
    \begin{subfigure}[t]{0.18\textwidth}
    	\includegraphics[scale=0.125]{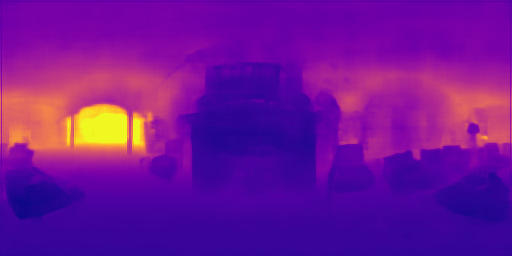}
    \end{subfigure}
    \begin{subfigure}[t]{0.18\textwidth}
    	\includegraphics[scale=0.125]{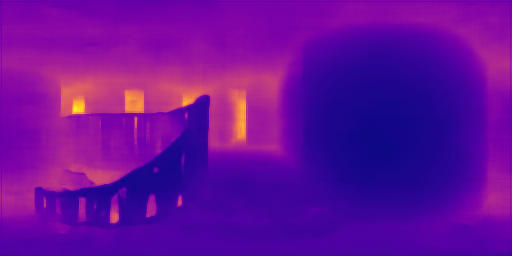}
    \end{subfigure}
    \caption{Qualitative results on our test split.}
    \label{fig:resAll}
\end{figure}
\begin{figure}[!t]
	{\tiny \hspace{2.95cm} "Indoors" split}{\tiny \hspace{4.2cm} "Room" split}
	
	\centering
	\begin{rotate}{90}{\tiny RGB}\end{rotate}
	\begin{subfigure}[t]{0.18\textwidth}
		\includegraphics[scale=0.125]{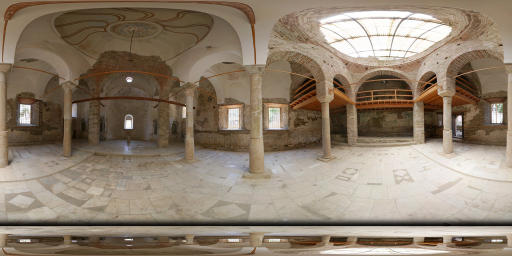}
	\end{subfigure}
	\begin{subfigure}[t]{0.18\textwidth}
		\includegraphics[scale=0.166]{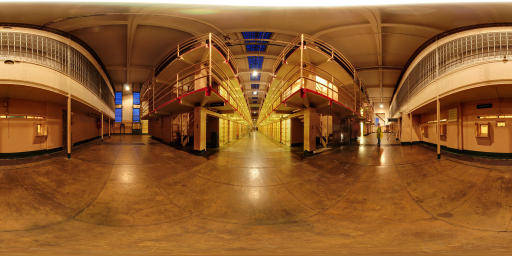}
	\end{subfigure}
	\begin{subfigure}[t]{0.18\textwidth}
		\includegraphics[scale=0.52]{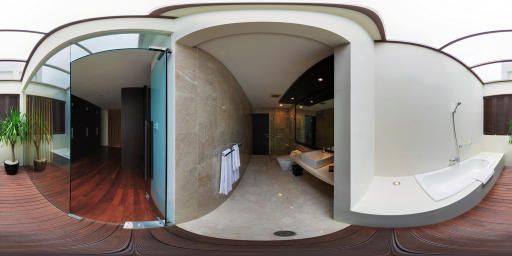}
	\end{subfigure}
	\begin{subfigure}[t]{0.18\textwidth}
		\includegraphics[scale=0.125]{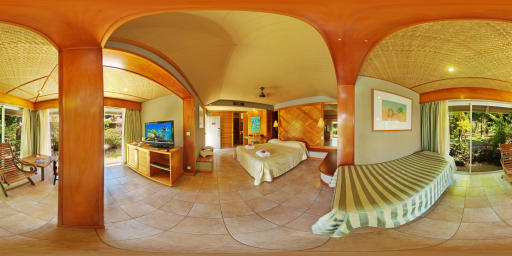}
	\end{subfigure}
	\begin{subfigure}[t]{0.18\textwidth}
		\includegraphics[scale=0.125]{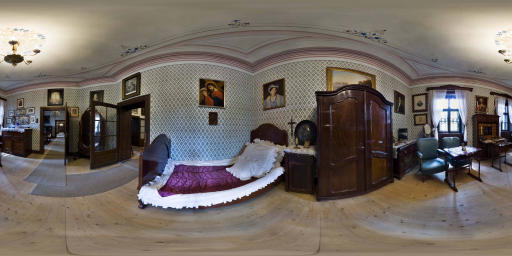}
	\end{subfigure}

	\begin{rotate}{90}{\scalebox{0.45}{Godard et al. \cite{monodepth17}}}\end{rotate}
	\begin{subfigure}[t]{0.18\textwidth}
		\includegraphics[scale=0.125]{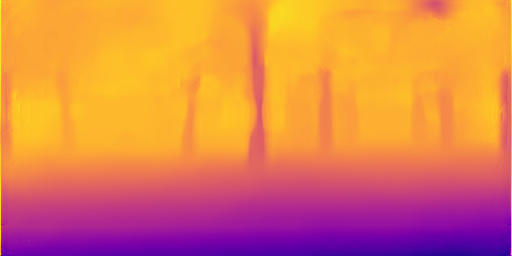}
	\end{subfigure}
	\begin{subfigure}[t]{0.18\textwidth}
		\includegraphics[scale=0.125]{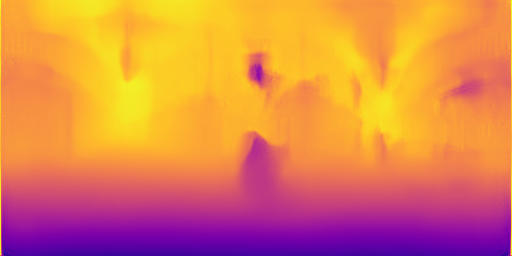}
	\end{subfigure}
	\begin{subfigure}[t]{0.18\textwidth}
		\includegraphics[scale=0.125]{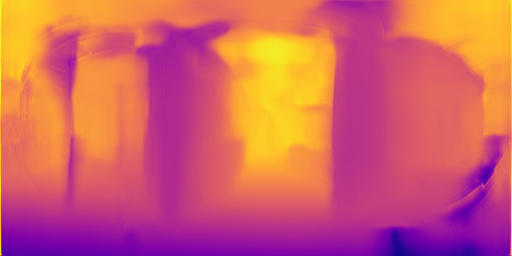}
	\end{subfigure}
	\begin{subfigure}[t]{0.18\textwidth}
		\includegraphics[scale=0.125]{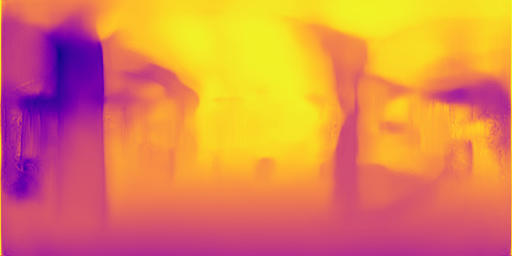}
	\end{subfigure}
	\begin{subfigure}[t]{0.18\textwidth}
		\includegraphics[scale=0.125]{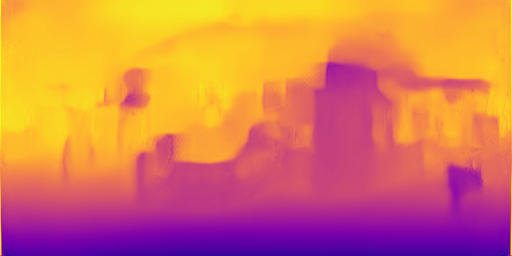}
	\end{subfigure}

	\begin{rotate}{90}{\scalebox{0.45}{Laina et al. \cite{laina2016deeper}}}\end{rotate}
	\begin{subfigure}[t]{0.18\textwidth}
		\includegraphics[scale=0.125]{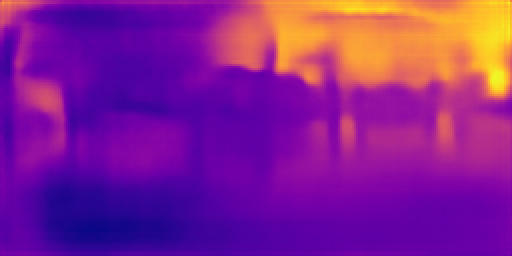}
	\end{subfigure}
	\begin{subfigure}[t]{0.18\textwidth}
		\includegraphics[scale=0.125]{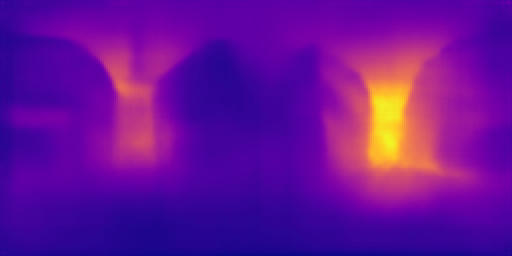}
	\end{subfigure}
	\begin{subfigure}[t]{0.18\textwidth}
		\includegraphics[scale=0.125]{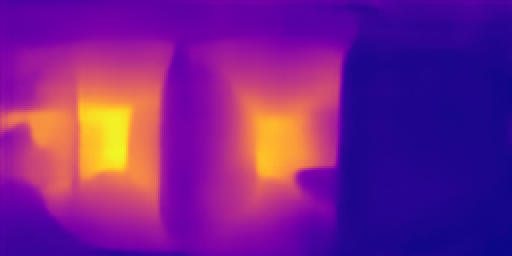}
	\end{subfigure}
	\begin{subfigure}[t]{0.18\textwidth}
		\includegraphics[scale=0.125]{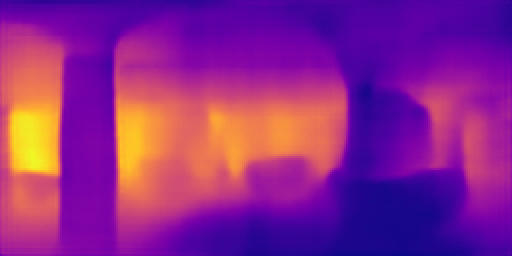}
	\end{subfigure}
	\begin{subfigure}[t]{0.18\textwidth}
		\includegraphics[scale=0.125]{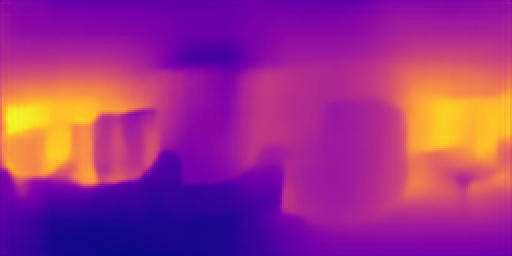}
	\end{subfigure}

	\begin{rotate}{90}{\scalebox{0.45}{Liu et al. \cite{liu2016learning}}}\end{rotate}
	\begin{subfigure}[t]{0.18\textwidth}
		\includegraphics[scale=0.125]{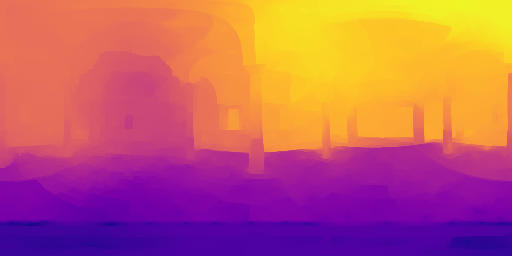}
	\end{subfigure}
	\begin{subfigure}[t]{0.18\textwidth}
		\includegraphics[scale=0.125]{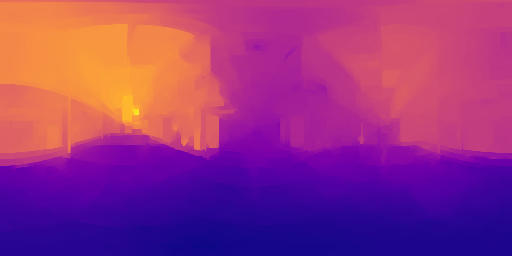}
	\end{subfigure}
	\begin{subfigure}[t]{0.18\textwidth}
		\includegraphics[scale=0.125]{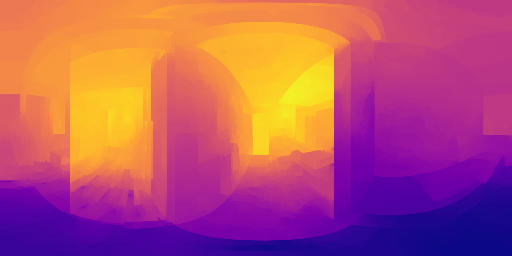}
	\end{subfigure}
	\begin{subfigure}[t]{0.18\textwidth}
		\includegraphics[scale=0.125]{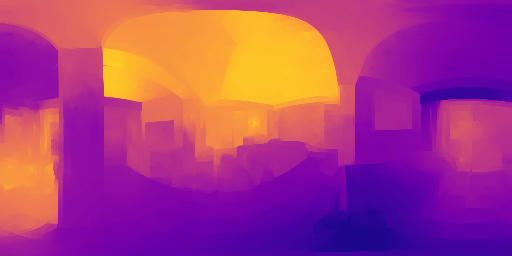}
	\end{subfigure}
	\begin{subfigure}[t]{0.18\textwidth}
		\includegraphics[scale=0.125]{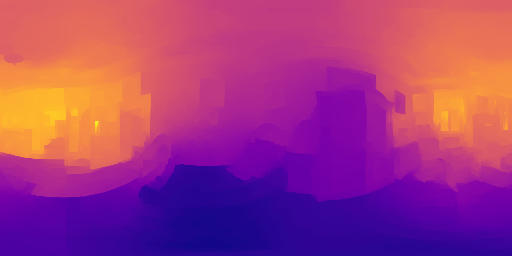}
	\end{subfigure}

	\begin{rotate}{90}{\scalebox{0.45}{Ours \unbal}}\end{rotate}
	\begin{subfigure}[t]{0.18\textwidth}
		\includegraphics[scale=0.125]{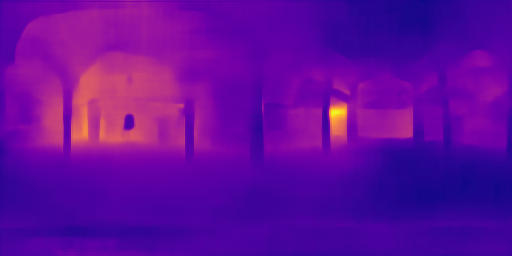}
	\end{subfigure}
	\begin{subfigure}[t]{0.18\textwidth}
		\includegraphics[scale=0.125]{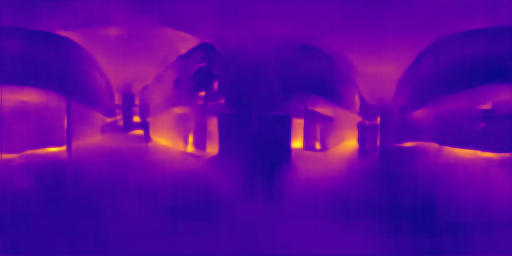}
	\end{subfigure}
	\begin{subfigure}[t]{0.18\textwidth}
		\includegraphics[scale=0.125]{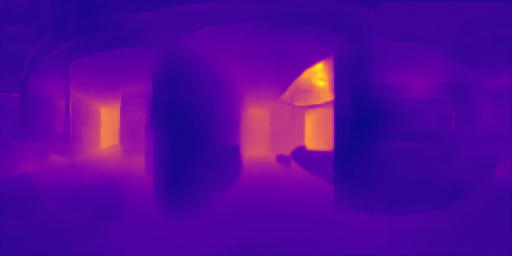}
	\end{subfigure}
	\begin{subfigure}[t]{0.18\textwidth}
		\includegraphics[scale=0.125]{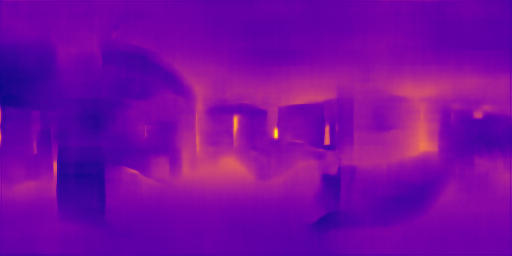}
	\end{subfigure}
	\begin{subfigure}[t]{0.18\textwidth}
		\includegraphics[scale=0.125]{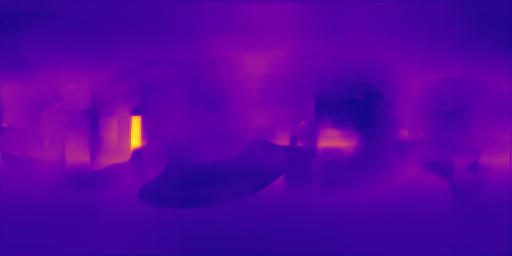}
	\end{subfigure}

	\begin{rotate}{90}{\scalebox{0.45}{Ours \rectdil}}\end{rotate}
	\begin{subfigure}[t]{0.18\textwidth}
		\includegraphics[scale=0.125]{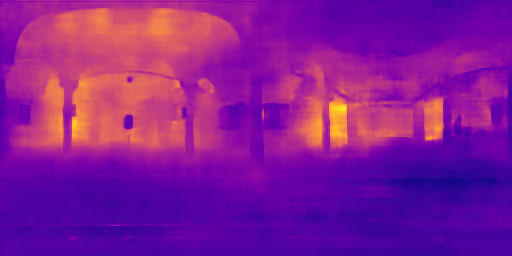}
	\end{subfigure}
	\begin{subfigure}[t]{0.18\textwidth}
		\includegraphics[scale=0.125]{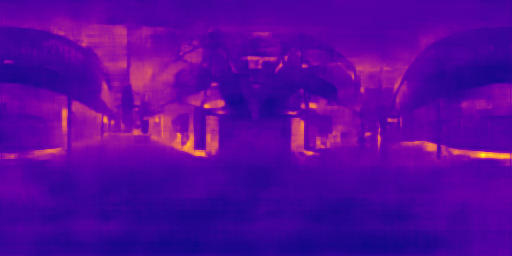}
	\end{subfigure}
	\begin{subfigure}[t]{0.18\textwidth}
		\includegraphics[scale=0.125]{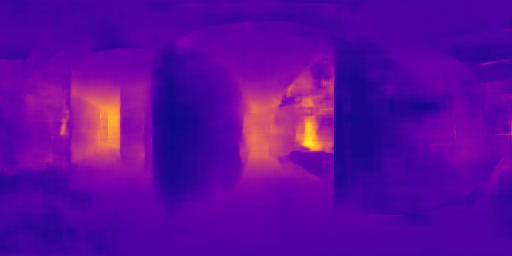}
	\end{subfigure}
	\begin{subfigure}[t]{0.18\textwidth}
		\includegraphics[scale=0.125]{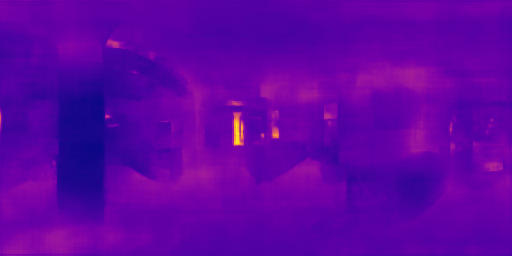}
	\end{subfigure}
	\begin{subfigure}[t]{0.18\textwidth}
		\includegraphics[scale=0.125]{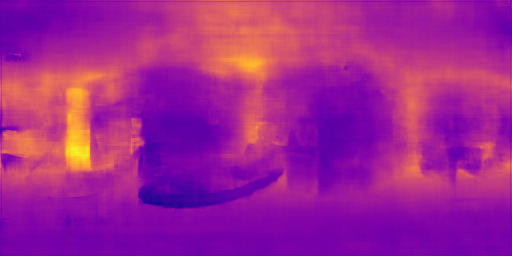}
	\end{subfigure}
	\caption{Qualitative results on the "Room" and "Indoors" Sun360 splits.}
	\label{fig:resQual}
\end{figure}

\textbf{Comparison against other methods}: Given that there are no other methods to perform dense depth estimation for \360 images, we assess its performance against the state of the art in monocular depth estimation models. Since the predictions of these methods are defined in different scales, we scale the estimated depth maps by a scalar \(\tilde{s}\), which matches their median with our ground truth like \cite{zhou2017unsupervised}, i.e. \(\tilde{s} = {median(D_{gt})} / {median(D_{pred})}\). Moreover, we evaluate the masked depth maps as mentioned in Section \ref{sec:datagen} in order to ignore the missing values. Table \ref{tab:resOthers} presents the results of of state-of-the-art methods when applied directly on our test split in the equirectangular domain. We offer results for the model of Laina et al. \cite{laina2016deeper}, trained with direct depth supervision in indoor scenes, Godard et al. \cite{monodepth17}, trained in an unsupervised manner in outdoor driving scenes using calibrated stereo pairs, and the method of Liu et al. \cite{liu2016learning}, which combines learning with CRFs and is trained in indoor scenes. As observed by the results, the performance of all the methods directly on equirectangular images is poor, and our main models outperform them. However, inferior performance is expected as these were not trained directly in the equirectangular domain but in perspective images. Nonetheless, Laina et al. \cite{laina2016deeper} and Liu et al. \cite{liu2016learning} achieve much better results than Godard et al. \cite{monodepth17}. This is also expected as the latter is trained in an outdoor setting, with very different statistics than our indoor dataset. 

\begin{table}[!t]
	\caption{Per cube face quantitative results against other monocular models.}
    \label{tab:resFaces}
	\centering
    \resizebox{\textwidth}{!}{
    \begin{tabular}{l | c c c c | c  c c  }
    	\hline 
   		Network & AbsRel \(\downarrow\)& SqRel \(\downarrow\)& RMSE \(\downarrow\)& RMSElog \(\downarrow\) & \(\delta < 1.25\) \(\uparrow\)& \(\delta < 1.25^2\) \(\uparrow\)& \(\delta N 1.25^3\) \(\uparrow\)\\
        \hline \hline
        \unbal 	& 0.0097 & 0.0062 & 0.1289 & 0.041 & 0.9245 & 0.9853 & 0.9955\\
        \rectdil  & \textbf{0.008} & \textbf{0.0042} & \textbf{0.1113} & \textbf{0.03504} & \textbf{0.9497} & \textbf{0.9907} & \textbf{0.9969}\\
        \hline
        Godard et al. \cite{monodepth17}  & 0.0453 & 0.1743 & 1.6559 & 0.1958 & 0.4524 & 0.7023 & 0.8315 \\
        Laina et al. \cite{laina2016deeper}  & 0.03 & 0.0549 & 0.3152 & 0.1033 & 0.6353 & 0.8616 & 0.9412\\
        Liu et al. \cite{liu2016learning}  & 0.0312 & 0.0532 & 0.3048 & 0.107 & 0.603 & 0.8412 & 0.9338\\
        \hline \hline
    \end{tabular}
   }
\end{table}
\begin{figure}[!t]
	\centering
	\includegraphics[scale = 0.38]{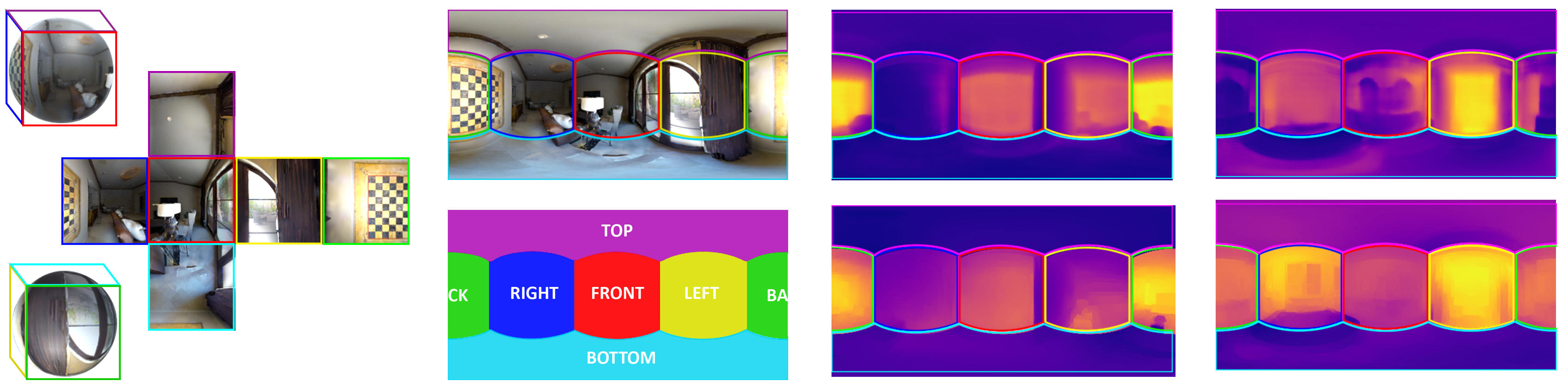}
    \caption{Cubemap projection (left) and merged monocular predictions (right).}
    \label{fig:resCubemap}
\end{figure}

For a more fair comparison we use a cubemap projection (Fig. \ref{fig:resCubemap} (left)) of all spherical images and then run each model on the projected cube faces which are typical perspective images. After acquiring the predictions, we merge all cube faces' depth maps by projecting them back to the equirectangular domain to be evaluated. However, since the top and bottom cube face projections will be mostly planar, we ignore them during evaluation of all metrics. While monocular performance is improved compared to when applied directly to equirectangular images, their quantitative performance is still inferior to our models. Further, the runtime performance is also worse as multiple inferences need to run, one for each face, incurring a much higher computational cost. Moreover, another apparent issue is the lack of consistency between the predictions of each face. This is shown in Fig. \ref{fig:resCubemap} (right) where it is clear that the depth scales of each face are different. This is in line with the observations in \cite{ruder2017artistic}, but is more pronounced in the depth estimation case, than the style transfer one. Based on this observation, we evaluate each cube face separately against the ground truth values of that face alone which is also median scaled separately. The average values of the front, back, right and left faces for each monocular model against the obtained by our models on the same faces alone are presented in Table \ref{tab:resFaces}. Although the performance of the monocular models is further improved, our models still perform better. This can be attributed to various reasons besides training directly on equirectangular domain. One explanation is that \360 images capture global information which can better help reasoning about relative depth and overall increase inference performance. The other is that our generated dataset is considerably larger and more diverse than other indoor datasets. In addition, the cube faces are projected out of $512 \times 256$ images and are thus, of lower quality/resolution than typical images these models were trained in.

\textbf{Qualitative Results: }To determine how well our models generalize, we examine their performance on completely unseen data found in the Sun360 dataset \cite{xiao2012recognizing}, where no ground truth depth is available. The Sun360 dataset comprises realistic environment captures and has also been used in the work of Yang et al. \cite{yang2016efficient} for room layout estimation. We offer some qualitative results on a data split from \cite{yang2016efficient}, referred to as "Room", as well as an additional split of indoor scenes that we select from the Sun360 dataset, referred to as "Indoors". These are presented in Fig. \ref{fig:resQual} for our two models as well as the monocular ones that were quantitatively evaluated. Our models are able to estimate the scenes' depth with the only monocular model to produce plausible results being the one of Laina et al. \cite{laina2016deeper}. We also observe that \unbal offers smoother predictions than the better performing \rectdilnospace, unlike the results obtained on our test split. More qualitative results can be found in the supplementary material where comparison with the method of Yang et al. \cite{yang2016efficient} is also offered.

\section{Conclusions}
\label{sec:conclusions}
We have presented a learning framework to estimate a scene's depth from a single \360 image. Our models were trained in a completely supervised manner with ground truth depth. To accomplish this, we overcame the dataset unavailability and difficulty in acquisition for paired \360 color and depth image pairs. This was achieved by re-using 3D datasets with both synthetic and real-world scanned indoors scenes and synthesizing a \360 dataset via rendering. \360 depth information can be useful for a variety of tasks, like in adding automation in the composition of 3D elements within spherical content \cite{rhee2017mr360}.
 
Since our approach is the first work for dense \360 depth estimation, there are many challenges that still need to be overcome. Our datasets only cover indoor cases, limiting the networks' applicability to outdoor settings, and are generated with perfect camera vertical alignment with constant lighting and no stitching artifacts. This issue is further accentuated as the scanned datasets had lighting information baked into them during scanning. This can potentially hamper robustness when applied in real world conditions that also contain a much higher dynamic range of luminosity.

For future work, we want to explore unsupervised learning approaches that are based on view synthesis as the supervisory signal. Furthermore, robustness to real world scenes can be achieved, either by utilizing GANs as generators of realistic content, or by using a discriminator to identify plausible/real images.
 
\textbf{Acknowledgements:} This work was supported and received funding from the European Union Horizon 2020 Framework Programme funded project Hyper360, under Grant Agreement no. 761934. We are also grateful and acknowledge the support of NVIDIA for a hardware donation.

\clearpage

\bibliographystyle{splncs}
\bibliography{egbib}

\end{document}